\pdfoutput=1

\documentclass[11pt]{article}

\usepackage{acl}

\usepackage{times}
\usepackage{latexsym}

\usepackage{multicol,multirow}

\usepackage[T1]{fontenc}

\usepackage[utf8]{inputenc}

\usepackage{microtype}

\usepackage{inconsolata}

\usepackage[shortlabels,inline]{enumitem}
\usepackage{cleveref}
\crefname{section}{s}{ss}
\crefname{section}{s}{ss}
\crefname{table}{Table}{Tables}
\crefname{figure}{Fig.}{Figs.}
\crefname{algorithm}{Alg.}{}
\crefname{ALC@unique}{Line}{Lines}
\crefname{equation}{Eq.}{Eqns.}
\crefname{appendix}{Appendix}{}
\crefformat{section}{\S#2#1#3}
\usepackage[normalem]{ulem}
\usepackage{graphicx}
\usepackage{caption}
\usepackage{subcaption}

\usepackage[type=none,unit=in]{fgruler}

\newcommand{\movedtoappendix}[1]{}
\title{CoRE: Condition-based Reasoning for Identifying Outcome Variance in Complex Events}

\author{ %
\textbf{Sai Vallurupalli, Francis Ferraro} \\
  Department of Computer Science and Electrical Engineering\\
  University of Maryland, Baltimore County\\
  Baltimore, MD 21250 USA \\
  \texttt{\{kolli,ferraro\}@umbc.edu} \\
  }

\begin{document}
\maketitle
 
\begin{abstract}
\label{sec:abstract}

Knowing which latent conditions lead to a particular outcome is useful for critically examining claims made about complex event outcomes.  Identifying implied conditions and examining their influence on an outcome is challenging. We handle this by combining and augmenting annotations from two existing datasets consisting of goals and states, and explore the influence of conditions through our research questions and \textit{Condition-based Reasoning} tasks. We examine open and closed LLMs of varying sizes and intent-alignment on our reasoning tasks and find that conditions are useful when not all context is available.  Models differ widely in their ability to generate and identify \textit{outcome-variant} conditions which affects their performance on outcome validation when conditions are used to replace missing context. Larger models like GPT-4o, are more cautious in such less constrained situations.

\end{abstract}

\section{Introduction}
\label{sec:intro}

\begin{figure}[t]
    \centering     
    \includegraphics[trim=0 0in 0 0.2in,  width=.48\textwidth]
    {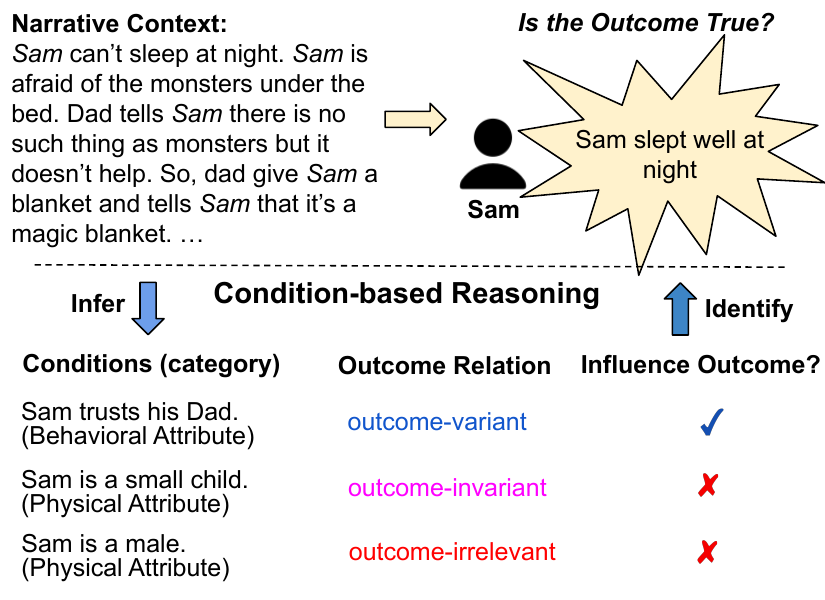}
	\caption{Conditions and their relationship to an outcome inspired by a story from PASTA/SAGA~\cite{ghosh-etal-2023-pasta,vallurupalli-etal-2024-saga}.  %
 }
	\label{fig:fig1}
 \vspace{-4mm}
\end{figure}

Knowing which conditions influence a goal's outcome is useful for understanding and planning goal-directed actions seen in complex events~\cite{CSIBRA200760}. %
Consider the following 5-sentence short story, also shown in \cref{fig:fig1}: 
\begin{quote}
\small
Sam can’t sleep at night. Sam is afraid of the monsters under the bed. Dad tells Sam there is no such thing as monsters but it doesn’t help. So, dad give Sam a blanket and tells Sam that it’s a magic blanket. Sam believes the blanket protects against monsters.
\end{quote}
While one can reasonably infer that Sam's goal (to sleep at night overcoming fears about monsters) was achieved \textit{in this story}, how stable is this outcome? What states, or conditions, relating to Sam or the situation help support the outcome, and if those were changed, would Sam's goal still be achieved? 
Some of these conditions, like  Sam trusting his/her dad or Sam being a small child, are relevant to the goal, with a high likelihood of influencing the outcome, while other conditions that might be true are irrelevant to the outcome---like Sam being male. 
Among relevant conditions, Sam trusting his/her dad directly influences the outcome and is \textbf{outcome-variant} because if Sam did not trust his/her dad the outcome is less likely to be true. Alternatively, Sam being a small child does not directly influence the outcome; it is \textbf{outcome-invariant} as the contrastive condition is unlikely to change the outcome. %
In this paper, we examine the interplay between state conditions and goal outcomes, and in particular, 
\textbf{%
we demonstrate how to use, and generate outcome-variant counterfactual conditions to effectively reason about whether a story outcome is true (or false) across altered narratives.} %

\begin{figure*}[t]
    \centering
    \begin{subfigure}[b]{0.32\textwidth}
    \includegraphics[trim=0in 0in 0in 0in, scale=.48 ]{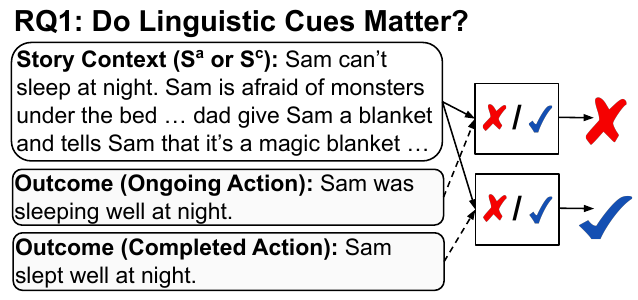}
    \caption{\label{fig:fig2a}We examine whether different descriptions, an ongoing Vs a completed action, change the truth value of an outcome. 
   }
    \end{subfigure}
    ~
    \begin{subfigure}[b]{0.32\textwidth}
    \includegraphics[trim=0in 0in 0in 0in, scale=.48]{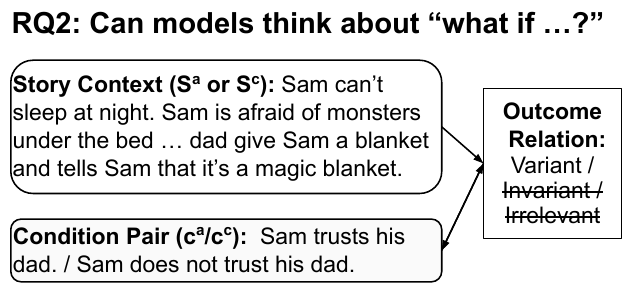}
    \caption{\label{fig:fig2b} We examine whether models are able to think counterfactually and generate/identify \textit{outcome-variant} conditions. }
    \end{subfigure}
    ~
    \begin{subfigure}[b]{0.32\textwidth}
    \includegraphics[trim=0in 0in 0in 0in, scale=.48]{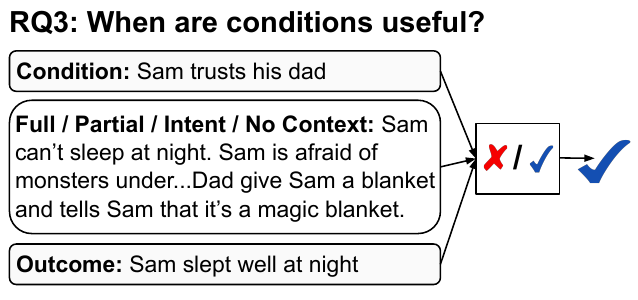}
    \caption{\label{fig:fig2c} We examine whether models are able to combine a condition with varying amounts of context to validate outcomes.}
    \end{subfigure}
     
    \caption{Our research questions explore outcome validation through our condition-based reasoning tasks.  We leverage PASTA~\cite{ghosh-etal-2023-pasta} and SAGA~\cite{vallurupalli-etal-2024-saga} datasets to generate outcomes of SAGA and News Stories and validate these using both our generated and PASTA's crowd-annotated conditions. }
    \label{fig:fig2}
 \vspace{-4mm}
\end{figure*}

This type of reasoning is challenging because: 
\begin{enumerate*}[(1)]
    \item Conditions that to 
    relate to entity properties and states are not always explicitly stated in a narrative but are implicitly understood through forming a coherent mental representation~\cite{ghosh-etal-2023-pasta} and acquiring this implicit knowledge is not easy.
    \item Outcomes can be vastly different even for slight variations in context~\cite{vallurupalli-etal-2024-saga} pointing to a need for identifying nuanced properties and states that have an influence on the outcome. 
    \item Knowing a condition's influence on the outcome is implicitly understood through constructing counterfactual mental representations~\cite{byrne,hugo} requiring robust means for acquiring counterfactual conditions and reasoning with them.
    \item Large Language Models (LLMs) ~\citep[\textit{inter alia}]{brown,ouyang2022traininglanguagemodelsfollow}, while powerful, do not necessarily perform well on tasks requiring counterfactual reasoning~\cite{fang-etal-2025-counterfactual,lin-2004-rouge,ghosh-etal-2023-pasta,qin-etal-2019-counterfactual}. 
    \item Linguistic differences influence action understanding~\cite{prus-etal-2024-human,zhou-etal-2023-navigating,salomon2013verb,hart2011learning} which can hinder an LLM in understanding condition and outcome descriptions.
\end{enumerate*}


Inspired by~\newcite{prus-etal-2024-human}'s findings that pragmatic knowledge influences how linguistic cues are interpreted, we examine how models use linguistic cues when understanding story outcomes.  We explore how implicit conditions influence outcomes and whether they can be used in place of missing context for validating story outcomes.  We leverage two previously released datasets PASTA~\cite{ghosh-etal-2023-pasta} and SAGA~\cite{vallurupalli-etal-2024-saga} consisting of participants' goal and state annotations (shown in \cref{fig:fig2}) and augment them for exploring conditions and outcomes. 
We address the following research questions and formulate our \textbf{Condition-based Reasoning} tasks to explore the capabilities of intent-aligned LLMs on these tasks.  We highlight the usefulness of conditions in determining story outcomes across a variety of alternative narratives, from short five sentence stories to real life newswire articles.

\textbf{RQ1:  Do Linguistic Cues Matter? } We explore models' ability to handle different types of outcome descriptions.  For the example in \cref{fig:fig1}, would models consider ``Sam was sleeping well'' to mean the same as ``Sam slept well?''  While these two descriptions do not mean the same according to Dowty's imperfective paradox~\cite{dowty},  \newcite{prus-etal-2024-human} show that meaning agreement is more complex.  We examine how models fare on such linguistic differences and find that the state-of-the-art models differ in performance on the imperfective paradox, but understand both ongoing and completed actions, with a slight preference for completed actions. See \cref{sec:RQ1} for more details.

\textbf{RQ2:  Can models think about ``what-if...?'' }  We examine whether models can reason counterfactually, through generating and identifying outcome-variant contrastive condition pairs.  We found that with one exception (Mistral-7B-Instruct-v0.3), all considered models are better at identifying whether a condition is outcome-related (which does not involve counterfactual thinking), but struggle at identifying outcome variance which requires counterfactual thinking. 
See \cref{sec:RQ2} for details.

\textbf{RQ3:  When are conditions useful? }  We examine if models are able to use conditions when context is unavailable or incomplete.  We find that while models are able to combine  conditions with available context to validate outcomes, some larger models 
are more cautious when the context is less constrained.  See \cref{sec:RQ3} for more details.

Overall, we find that performance on outcome validation improves by 2-6\% (macro F1 score) when  \textit{outcome-variant} (as opposed to \textit{outcome-invariant}) conditions are used with the story context.  Since identifying or generating \textit{outcome-variant} conditions is not easy, performance drops by 3-6\% for generated conditions when compared to annotated conditions. 
For news stories, performance drops by up to 2-18\%, when compared to the PASTA story generated conditions, likely attributable to a combination of the domain change and models' going beyond the provided context and using their knowledge of past news events from their pretraining.  See \cref{sec:news} for more details. 

Our contributions in this work include introducing the concept of conditions and their relationships to an outcome. We artfully combine and bridge two existing datasets to explore intertwined aspects of conditions and outcomes.  Through our research questions  we examine how LLMs leverage conditions for outcome validation and compare their performance on the practical task of validating outcomes of News stories. %
Our data and code are available at https://github.com/saiumbc/CoRE. %

\section{Related Work}
\label{sec:related}

Examining the influence of states and actions on goal outcomes is an active research area in the cross-disciplinary fields of cognitive science and psychology ~\cite{hommel,custers,niv,amir2024states}. The utility of counterfactual conditionals in reasoning~\cite{goodman,filho}  has a long history in linguistics \& psychology~\cite{byrne1999deductive, byrne2019counterfactuals}.  

\paragraph{States, Actions \& Counterfactuals:}
The hypotheses in abductive natural language inference ($\alpha$-NLI)  task~\cite{chandra2020a} and defeasible inference ($\delta$-NLI)~\cite{rudinger-etal-2020-thinking} are similar to the conditions we examine, however, these do not reference any specific goal or outcome.  $\delta$-NLI  WIQA~\cite{tandon-etal-2019-wiqa} examined counterfactual situations through preturbing states with ``what if'' type of questions and obtained changes in action outcomes. PASTA~\cite{ghosh-etal-2023-pasta} examined implied states and preturbed these states to examine changes in situational narratives.  We condition PASTA states on goal achievement and obtain state conditions that influence goal outcomes. 

\paragraph{Goals \& Outcomes:}
 Goal oriented reasoning has been explored in participant narratives~\cite{rahimtoroghi-etal-2017-modelling}, news actions~\cite{jiang-riloff-2018-learning} and procedural text~\cite{zhang-etal-2020-reasoning}.  LLM's goal reasoning has been explored using participant goals by \newcite{bellos-etal-2024-large} and SAGA~\cite{vallurupalli-etal-2024-saga}. We extend SAGA to examine conditions' influence on goal outcomes.  

 \paragraph{Linguistic Understanding:}
 Studying aspect for temporal reasoning has a long history in computational linguistics\cite{moens-steedman-1988-temporal,siegel-mckeown-2000-learning}. More recent work focused on aspect to study verb classes in text and their affect on textual entailment~\cite{kober-etal-2020-aspectuality} and cross-domain data utility\cite{alikhani-stone-2019-caption}. Inspired by \cite{friedrich-etal-2023-kind} and \cite{prus-etal-2024-human} we examine LLMs' aspectual understanding for identifying goal outcomes.  
\section{Data Annotations }
\label{sec:pasta-saga}
In the following paragraphs, we briefly describe the PASTA~\cite{ghosh-etal-2023-pasta} and SAGA~\cite{vallurupalli-etal-2024-saga} datasets and how these datasets' annotations relate to conditions and outcomes.  

\paragraph{(Counterfactual) State Condition of Participants in Narratives:} The PASTA dataset is a collection of stories and implied states supported by the stories.  For a given 5-sentence ROC Story~\cite{mostafazadeh-etal-2016-story}, crowd workers infer and describe a state implied by one or more of the sentences in the story (the annotations identify these sentences). For this state they also describe a perturbed state (a ``what-if'' type of counterfactual state) that is  unsupported by the original story; they also minimally alter the story to support the perturbed state ensuring it does not support the original state. The dataset contains 5028 original stories and up to 3 state annotations per story, leading to 5715 state pairs and 5715 alternate stories.  See \cref{tab:pasta-data-samples} for examples of data annotations.

\paragraph{Goal Outcomes of Participants in Narratives:}
The SAGA dataset is a collection of goal related annotations for a subset of the PASTA story collection.  Crowd workers describe an overarching goal of a volitional participant in the original story, how each actual story sentence relates to the goal and whether the goal is achieved;  the described goal is what the participant hopes to achieve through their actions in the story and beyond.  Alternate stories corresponding to the original story are examined to assess whether the actions in these stories can achieve the goal.   The dataset contains three goal annotations for each participant in a story, for up to 4 story participants obtaining a total of 2785 goal annotations for 886 original stories (449 of these have 951 corresponding alternate stories).  See \cref{tab:saga-data-samples} for examples of data annotations.  We describe only the annotations we use in this work. For the complete list, please refer to \newcite{vallurupalli-etal-2024-saga}. 

\paragraph{Conditions and Outcomes in Alternative Narratives: } 
SAGA goals allow the examination of an entity's goal achievement in alternate stories and PASTA states\footnote{We refer to PASTA states as conditions in this work.} describe various properties and states of an entity that are supported by at least one of the alternate stories.  Studying the interrelationships between these, we find that: 
\begin{enumerate*}[(a)]
\item several different types of conditions can be inferred from a narrative,  
\item not all conditions are related to a goal (these are \textit{outcome-irrelevant}), and 
\item slight changes in a narrative can lead to a contrastive condition but not all contrastive condition pairs lead to a different goal outcome.  
\end{enumerate*}
In some pairs, one state leads to achievement and the other not (we consider these to be \textit{outcome-variant}) and in some pairs, both states lead either to goal achievement or not (we consider these to be \textit{outcome-invariant}).  See \cref{sec:conditions-outcomes} for a detailed description with examples.  

With this knowledge we augment PASTA and SAGA annotations  for exploring conditions and outcomes.  We consider an outcome to be a statement indicating the achievement (or not) of a participant's goal and want to identify whether the outcome is true for the story.  For example in \cref{fig:fig1}, the outcome is a statement about Sam achieving the goal of ``sleeping well at night'' and it is true for the story.  We use a combination of prompting GPT3.5-Turbo and automatic methods to derive all annotations we need from PASTA/SAGA except for the outcome label. Our evaluation of conditions is based on outcome labels and annotating these requires understanding the story and deep reasoning. Hence we use an expert to obtain these label annotations.  We describe our annotation process in \cref{app:data_augmentation} listing the annotation statistics in \cref{apptab:annotation-details} and data examples in \cref{tab:core-data-samples}.

\section{Conditions \& Outcomes}
\label{sec:conditions-outcomes}
\subsection{Conditions}  Conditions inferred from a story include properties of entities ranging from the intrinsic (inherent attributes such as \textit{Rob was immature}, \textit{Cindy had long hair}, etc.) to the extrinsic (attributes resulting from other entities such as \textit{Timothy has an old TV}, \textit{Janice's closet is messy}, etc.).  We expand the 3 categories used in PASTA \cite{ghosh-etal-2023-pasta}, for error analysis on 200 random states on the story state inference task, to 4 categories and group all  conditions into these as follows: 
\begin{enumerate}[(a),itemsep=0pt,leftmargin=*]
\item \textbf{Physical:} This category includes natural physical attributes of an entity such as size, age, place etc., such as \textit{Cindy had long hair}, \textit{The cake is really big}, \textit{The sky was cloudy etc}. 

\item \textbf{Functional:} This category includes attributes that influence an entity's capability to perform actions, e.g., \textit{The machine was jammed}, \textit{Timothy has an old TV}, etc.

\item \textbf{Knowledge:} This category includes mental states of knowing information (knowledge about self or other entities, or pragmatic world knowledge), e.g., \textit{Lisa knows her family well}, \textit{Nancy is up to date with technology}, \textit{Bill was aware that the cap had been loosened}, etc.

\item \textbf{Behavioral:} This category includes behavioral aspects of an entity for example: \textit{Cindy's dog is a biter}, \textit{Charlie loves candies}, \textit{the librarian is responsible}, \textit{Amber is frugal}, etc.
\end{enumerate}
\noindent This grouping identifies categories that are likely to be outcome-variant and useful for reasoning.

\subsection{Outcome Relationships} The relationship between a condition and an outcome can be understood through counterfactual reasoning. Consider a pair of alternate story\footnote{We use 
 “narrative” and “story” interchangeably, even though we acknowledge they have important differences.} contexts $S_a$ and $S_c$ that support a contrastive condition pair $c_a$ (an actual condition) and $c_c$ (a counterfactual condition) respectively where both stories contain goal-oriented events to achieve a common goal. As an outcome description $O$ can be either true or false for a given story, we define 3 types of relationships:  

A condition is \textbf{outcome-variant}, when $O$ for $S_a$ and $S_c$ are different with one being true and the other being false.  For the example in \cref{fig:fig1}, we consider \textit{Sam trusts his Dad} as $c_a$ and \textit{Sam does not trust his dad} as $c_c$.  The outcome \textit{Sam slept well at night} is true for $S_a$ but false for $S_c$.  Hence, both conditions are outcome-variant.

A condition is \textbf{outcome-invariant}, when outcome $O$ is true (or false) for both $S_a$ and $S_c$.  For the example in \cref{fig:fig1},  \textit{Sam is a small kid} is the $c_a$ and \textit{Sam is a big kid} the $c_c$. Both conditions are outcome-invariant because the outcome \textit{Sam slept well at night} is true for both contexts, $S_a$ and $S_c$.  While these conditions can be seen as outcome-variant through the use of commonsense inference such as \textit{a small kid trusts their parent}, in such cases, the commonsense inferred condition will be the outcome-variant condition.

A condition is \textbf{outcome-irrelevant} if it and its counterfactual condition have no bearing on the outcome.  For the example in \cref{fig:fig1}, the actual condition \textit{Sam is a boy} and the corresponding counterfactual condition \textit{Sam is a girl} have no relevance to the outcome \textit{Sam slept well at night}.

\subsection{Lexical Cues \& Imperfective Paradox}
\label{sec:cues-paradox}
Our outcomes describe the end-result of volitional actions.  In \cref{fig:fig1} Sam slept well is the outcome of several goal-oriented actions.  This outcome can be interpreted as completed or ongoing (incomplete) based on linguistic cues such as verb aspect and world knowledge ~\cite{GIVÓN,magliano2000verb,madden2003does} and annotated as such. Whether an outcome is true depends on accurately deciphering whether the planned goal is achieved, regardless of linguistic differences.  In \cref{fig:fig1},
whether the outcome is described as Sam slept well or Sam was sleeping well, we want to know whether Sam achieved the goal ``of sleeping at night.''

\paragraph{Imperfective Paradox:} \Citet{dowty} notes that according to the ``Imperfective Paradox,'' an accomplishment described as a past ongoing action is not necessarily the same as a past completed action, i.e., that Sam slept well is not the same as Sam was sleeping well (we consider \textit{sleeping well} and \textit{slept well} in the context of Sam's goal accomplishment).  

\paragraph{Usefulness of Aspect:}
The semantic property of lexical aspect associated with verbs 
helps us understand how actions unravel over time~\cite{smith83,vendler}. Grammatical aspect helps us distinguish between a completed  and an ongoing action through analyzing different view-points--of the entire situation vs. a part of it~\cite{smith1999}. Lexical aspect helps us understand \textit{sleep} as an activity and grammatical aspect helps us consider when \textit{slept} and \textit{sleeping} mean the same.  (See ~\newcite{friedrich-etal-2023-kind} for a more detailed discussion).

\paragraph{Our study:} \citet{prus-etal-2024-human} show that pragmatic reasoning rooted in world knowledge influences how aspect is understood and that an LLM's pragmatic world knowledge (acquired during pretraining) leads to a different aspectual understanding than that of humans.  We extend their study to examine to what extent LLMs' aspectual knowledge affects condition and outcome understanding, using the imperfective Paradox with and without story context. Specifically, we examine if LLMs consider `Sam was sleeping well' to be different from `Sam slept well' for determining if Sam's goal is achieved for several alternate story contexts and when no context is provided.  We examine whether models can leverage the provided context and perform a pragmatic view-point analysis that is dictated by the context to improve their inferences. 

\section{Outcome Validation through RQs}
\label{sec:Experiments} 


We examine whether an outcome is possible for a condition known about an entity through our research questions.  We use both crowd-annotated and LLM-generated conditions, where not every condition is supported by the story context or related to the outcome.  We formulate reasoning tasks with the aim to compare various models' performance on identifying and generating outcome-variant conditions.  We examine the conditions' usefulness in  validating outcomes of SAGA and News Stories.   

\paragraph{Models:} We examine well-known closed and open, human intent-aligned LLMs of different sizes: GPT-4o-mini, GPT-4o, FlanT5-XXL, LLaMA-3.1-8B-Instruct (Llama-8BI), Mistral-7B-Instruct-v0.3 (Mistral-7BI) and LLaMA-3.1-70B-Instruct (Llama-70BI).  We refer to the models by the shortened names in parentheses.  We examine the impact of model type, size and the alignment.

\paragraph{Inference:}  For the tasks examined in RQ1 and RQ3, we use zero-shot prompting to examine models' inherent knowledge. For the tasks examined in RQ2 we use few-shot prompting which is required as models do not inherently understand how to generate or identify outcome-variance. Inference cost for GPT models was less than \$100.%

\subsection{RQ1:  Do Linguistic Cues Matter?}
\label{sec:RQ1} 
 We examine aspect understanding using two sets of outcome descriptions (past ongoing \& past completed actions) using 
 \begin{enumerate*}[(a)]
 \item  Dowty's imperfective paradox (discussed in \cref{sec:cues-paradox}) and 
 \item direct questioning. See prompts for both tasks in \cref{tab:rq1-prompts}. 
\end{enumerate*}

We compare performance on the imperfective paradox with and without the story context to examine whether models are able to improve upon their aspectual understanding leveraging the provided context.   We extend the prompt from \citet{prus-etal-2024-human} to
\begin{enumerate*}[(i)] 
\item include an \textit{Unsure} option, in addition to \textit{Yes} and \textit{No}, which is useful when stories do not provide enough information relating to an outcome, and 
\item optionally provide story context.
\end{enumerate*}  

With direct questioning, we examine if models have a preference for a specific type of linguistic cue and  if their understanding of the two descriptions agrees with that of human preference.  For the latter, we  compute Cohen's Kappa ($\kappa$) between the labels for both descriptions.  
 
\paragraph{Evaluation:}  We compare the F1 values for the 3 labels (True/False/Unsure) of the imperfective paradox without and with story context.  We derive the gold labels for this task as follows: we use the gold label from the on-going action, when the label for the completed action is true, otherwise, we use the label for the completed action.  
For direct questioning, we compare F1 values for all 3 labels prompting twice for the two outcome descriptions.   We compute Cohen's Kappa ($\kappa$) between model generated labels and gold labels for both prompts. 

\begin{table}[t]
 \centering
\resizebox{.98\columnwidth}{!}{
 \begin{tabular}{|l|c|c|}
 \hline
Model&  No Context & Story Context \\ 
\hline
Flan-T5-XXL  & .68/.00/.09 & .71/.21/.00 \\
GPT-4o-mini & .64/.27/.00 & .73/.49/.02 \\
GPT-4o      & .37/.50/.15 & .69/.64/.11 \\
Mistral-7BI & .69/.08/.00 & .47/.61/.05 \\
Llama-8BI  & .48/.44/.00 & .23/.53/.00 \\
Llama-70BI  &  .41/.46/.00&  .44/.56/.00\\
 \hline 
\end{tabular}
}
 \caption{F1 scores for True/False/Unsure answers to the Dowty's Imperfective Paradox with (`Story Context' column) and without (`No Context' column) the story context.  Access to context improves Flan-T5 \& GPT.}
\label{tab:aspect}
\vspace{-5mm}
\end{table}

\paragraph{Results \& Discussion:}
According to results from the imperfective paradox (see \cref{tab:aspect}),  FlanT5-XXL, GPT-4o-mini and Mistral-7BI are better than GPT-4o and Llama models at aspectual understanding. 
FlanT5-XXL, GPT-4o-mini and GPT-4o are able leverage context to improve upon their base aspectual understanding with GPT-4o improving by a large amount;  Mistral 7BI and the Llama models are unable and perform poorly. This performance indicates how models handle linguistic complexities in our tasks (see examples in \cref{tab:rq1-examples}).

The results from direct questioning (in \cref{tab:linguistic-cues}) show that all models except GPT-4o have a slight preference towards completed action descriptions with their better performance.  This is true for Llama and Mistral models as well despite their poor performance on the imperfective paradox shown in \cref{tab:aspect}.  The high agreement between in-progress and completed descriptions, for all models except GPT-4o, is comparable to the agreement seen in gold labels (within $\pm$ .06).  Some models like FlanT5-XXL do not distinguish the linguistic nuance between the two descriptions 
and others like the Llama and GPT models distinguish more. GPT-4o is more cautious and generates an `Unsure' label in nuanced situations.  See examples in \cref{tab:rq1-examples-more}.

\begin{table}[t]
 \centering
\resizebox{.98\columnwidth}{!}{
 \begin{tabular}{|l|c|c|c|}
 \hline
Model& In-progress & Completed  &   Cohen's  \\ 
& Action & Action  & kappa $\kappa  $ \\ 
\hline
Gold Label & - & - &.77\\
FlanT5-XXL & .81/.67/.12 & .82/.71/.19 &.84\\
GPT-4o-mini& .80/.72/.11 & .81/.77/.24 & .74\\
GPT-4o      & .80/.79/.46 & .80/.79/.46 & .63\\
Mistral-7BI& .75/.68/.02 & .76/.70/.10 & .77 \\
Llama-8BI  & .77/.68/.00 & .79/.72/.00 & .71\\
Llama-70BI & .75/.69/.05 &  .80/.74/.19 & .72 \\
 \hline 
\end{tabular}
}
 \caption{F1 values for models output of True/False/ Unsure with directly questioning if an outcome description is true for a story context and the two sets of outcome descriptions. The $\kappa$ column shows agreement between labels generated for both descriptions (and gold labels). }
\label{tab:linguistic-cues}
\vspace{-5mm}
\end{table}

\subsection{RQ2: Can Models Think ``What If ...?'' } 
\label{sec:RQ2} 
We examine whether models can think counterfactually, when generating and identifying conditions, formulating both generation and classification tasks as follows: 
\begin{enumerate*}[(a)] 
\item We examine if models can generate contrastive condition pairs that are \textit{outcome-variant} (when prompted with similar examples) and  \textit{outcome-invariant} (again when prompted with similar examples) 
\item  We examine whether models can identify a condition's outcome-relevance and a contrastive pair's outcome variance.
\end{enumerate*}

\subsubsection{Generating Outcome Relationships}
\label{sec:RQ2-generation} 
We prompt models with a story and an outcome and generate contrastive condition pairs where the first condition is expected to be supported by the story and the second condition is expected to be a counterfactual condition (see \cref{tab:rq2-prompts} for the prompt).  To ensure, the conditions are contrastive, we use incontext learning examples.

\begin{table*}[t]
\centering
\resizebox{.98\textwidth}{!}{
\small
\begin{tabular}[trim=0 .2in 0 0,  width=.98\linewidth]{p{0.43\linewidth} | p{0.56\linewidth} }
\hline\hline
Prompt for generating a condition pair & Model \& Generated condition pair   \\ 
\hline\hline
\textbf{A:}  Sam can't sleep at night. He is afraid of the monsters under the bed. Dad tells him there is no such thing as monsters but it doesn't help. So Dad gives him a blanket and tells him that it's a magic blanket. Sam believes that he is protected by the blanket and he sleeps well. \newpage
\textbf{B:} Sam had a restful night's sleep.
\newpage Generate a pair of contrastive conditions relevant to the context in A and the statement in B.  Please make sure the conditions are stative expressions and do not describe activities.
&answer: Sam didn't believe that the blanket was a magic blanket. [FlanT5-XXL] \newpage Sam feels anxious about sleeping., Sam feels calm about sleeping. [GPT-4o-mini] \newpage
Sam's belief in the magic of the blanket is strong.,Sam's belief in the magic of the blanket is weak. [GPT-4o] \newpage
Sam is afraid of monsters under the bed., Sam had a peaceful sleep. [Mistral-7BI] \newpage Sam feels protected from the monsters, Sam feels vulnerable to the monsters [Llama-8BI] \newpage
Sam feels unsafe at night, Sam feels safe at night [Llama-70BI] \\

\hline\hline
\end{tabular}
}
\caption{Examples of Model Generated Conditions.  Unlike other models which generate fairly good condition pairs, FlanT5-XXL does not generate a pair and Mistral-7BI does not generate a contrastive and variant pair.  }
\label{tab:rq2-generated-conditions}
\vspace{-3mm}

\end{table*}

\begin{table}[t]
 \centering
 \begin{tabular}{|l|c|c|c|c|}
 \hline
Model&Con. & Rel. & Sup. & Var.\\
\hline
FlanT5-XXL & 10& 88& 66&  6 \\
GPT-4o-mini& 96&100& 80&100 \\
GPT-4o     & 91& 97& 82& 94 \\
Mistral-7BI& 94&100& 94& 88 \\
Llama-8BI  & 94&100& 86& 94 \\
Llama-70BI & 100&100&96& 100\\
\hline
\end{tabular}
 \caption{Evaluation of generated condition pairs on 4 factors using 100 random generations from each model. Llama-70BI generates condition pairs of high quality. }
\label{tab:condition-generation}
\vspace{-5mm}
\end{table}

\paragraph{Evaluation:}    We manually examine 100 random condition pairs selecting an equal number of original and alternate stories and an equal number of pairs expected to be outcome variant and invariant. Our evaluation rubric is designed to obtain a binary answer of Yes or No and consists of the following questions.   
\begin{enumerate*}[(i),topsep=-5px,topsep=0px,noitemsep]
\item \textit{Con.}: Are conditions in the pair contrastive? 
\item \textit{Rel.}: Is least one of the conditions outcome relevant?
\item \textit{Sup.}: Is the first condition in the pair supported by the story?
\item \textit{Var.}: Is the pair outcome-variant for the story context?
\end{enumerate*}
When a single condition in the pair is outcome relevant, it should be supported for the pair to be relevant.

\paragraph{Results \& Discussion:} 
We show an example prompt and the conditions generated by the various models in \cref{tab:rq2-generated-conditions}.  Generating contrastive and supported conditions that are outcome- relevant and variant indicate a model's counterfactual thinking and have an impact on outcome validation (See \cref{tab:rq2-generated-conditions-more} for erroneous generations).   Using exclusively \textit{outcome-variant} or \textit{outcome-invariant} did not ensure the generation of condition pairs of the same type.  Our manual evaluation results (in \cref{tab:condition-generation})  show that models are unable to distinguish between the two types and always generate a mix of mostly  outcome-variant conditions.  The pairs are mostly contrastive ($< 10\%$ are not) and outcome-relevant but are not always ordered (the first condition being the one supported by the story and the second being a counterfactual). All models, except, FlanT5-XXL and Llama70bI, are better at generating a pair that is ordered or outcome-variant but   Llama70BI is better at both.  FlanT5-XXL is poor in all evaluated criteria, however, the generated condition contains information highly useful for outcome  validation (see \cref{sec:RQ3} results).

\subsubsection{Identifying Outcome Relationships}
\label{sec:RQ2-identification} 
To identify a condition pair's outcome variance, we examine both a single-step reasoning and a multi-step reasoning where we using a standard Chain-of-Thought (CoT)~\cite{wei} approach (use gold labels for each condition's relevance and the pair's variance to reason step by step in the in-context example with the model performing the same during inference) and a variant approach where each condition's outcome relevance from two additional prompts is provided at inference time.   %
See \cref{tab:rq2-prompts} for these prompts.

\begin{table}[t]
 \centering
\resizebox{.98\columnwidth}{!}{
 \begin{tabular}{|l|c|c|c|c|}
 \hline
Model& Single-step & Stand. CoT & Alter. CoT \\ 
\hline
FlanT5-XXL &.43/.55& .46/.56 & .38/.62  \\
GPT-4o-mini& .54/.38  & .50/.47 &  .42/.59   \\
GPT-4o     &.52/.59 & .52/.62&.41/.71  \\
Mistral-7BI& .32/.67& .00/.72& .09/.72 \\
Llama-8BI  & .53/.35 & .38/.60& .55/.24 \\
Llama-70BI & .57/.37 & .54/.36& .54/.48 \\
 \hline 
\end{tabular}
}
 \caption{Identifying a condition pair's outcome-variance is difficult for models.  We compare F1 scores for whether a condition pair is outcome- variant/invariant using single and multi-step reasoning. Standard CoT uses gold labels for in the in-context examples; alternative CoT uses answers from condition-relevance (see \cref{tab:outcome-relevance}) during inference. }
\label{tab:outcome-variance}
\end{table}

\begin{table}[t]
 \centering
\resizebox{.98\columnwidth}{!}{
 \begin{tabular}{|l|c|c|c|c|}
 \hline
Model&Irrelevant & Relevant & Support &Oppose\\ 
\hline
FlanT5-XXL & .02& .87& .55 &.01 \\
GPT-4o-mini& .02& .86& .44 & .57\\ 
GPT-4o     & .28& .78&  .32 & .54\\
Mistral-7BI& .37& .04& .05 & .00\\
Llama-8BI  & .00& .87& .42 & .55  \\
Llama-70BI & .00& .87& .48 & .59\\
\hline
\end{tabular}
}
 \caption{Identifying a condition's relevance to the outcome is easier for models but they struggle to identify whether a condition supports or opposes the outcome. Identifying whether a pair is variant depends on the support/oppose relation from the intermediate step of Alternate CoT lowering its performance. }
\label{tab:outcome-relevance}
\vspace{-5mm}
\end{table}

\paragraph{Evaluation:}
We compare performance of the single-step and multi-step approaches using the F1 values for each task label. We interpret the 3 labels of True/False/Unsure for binary outcome- relevance and variance as follows: True and False labels are relevant (i.e.,  supporting \& opposing relations) and unsure labels are irrelevant. For variance, True labels are variant, False \& Unsure labels are invariant. We consider these groupings appropriate for positively identifying variance and relevance.  

\paragraph{Results \& Discussion:} 
Models find it difficult to identify  outcome- variant and invariant conditions, as seen by the lower F1 scores for both labels (see  \cref{tab:outcome-variance}). GPT-4o and Llama-70BI are slightly better than the smaller models with Mistral-7BI being the worst.  The standard CoT prompting improves only FlanT5-XXL with other models only improving in identifying invariance. Performance for the alternative CoT prompt suffers from relying on an intermediate step to identify a condition's support.
Our results (in \cref{tab:outcome-relevance}) show that all models except Mistral-7BI are good at identifying whether a condition is relevant but are poor at identifying whether it supports or opposes the outcome.  This is likely due to the nature of our alternate stories where minimal changes can alter the outcome.

\begin{figure}[t] 
    \centering  
    \vspace{-6mm}
    \begin{subfigure}[b]{0.49\columnwidth}
    \vspace{-3mm}
    \includegraphics[trim=2in 0in 0in 0in, scale=.45]{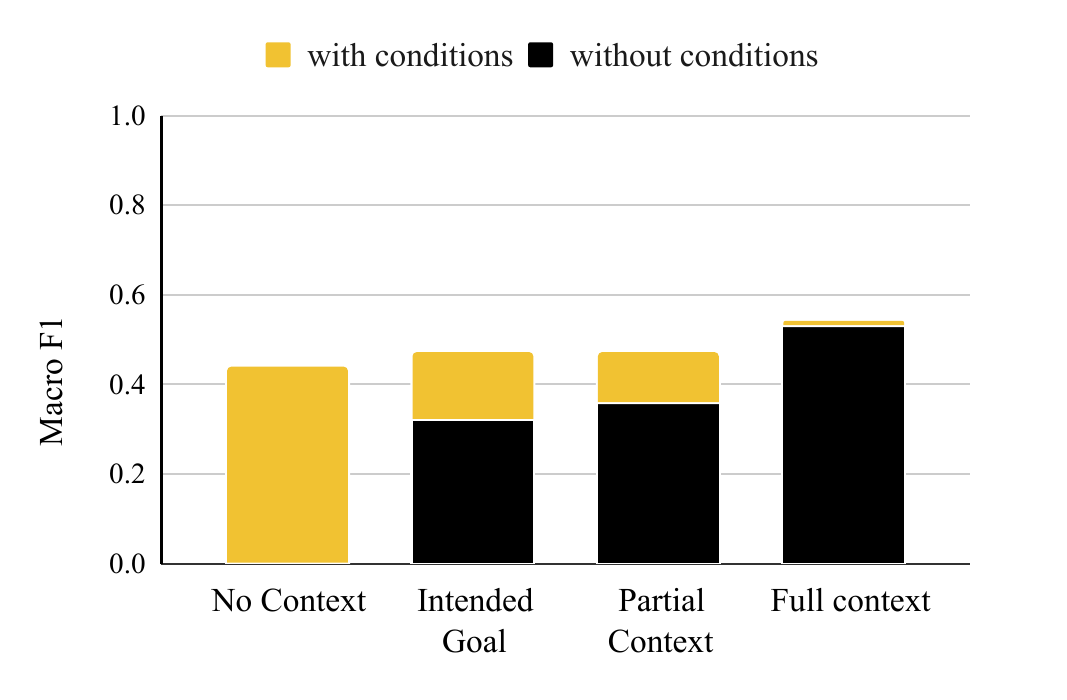}
    \vspace{-8mm}
	\caption{ \textbf{Llama-8BI} leverages conditions to improve performance.}
    \end{subfigure}
    ~
    \begin{subfigure}[b]{0.49\columnwidth}
    \vspace{-4mm}
    \includegraphics[trim=2in 0in 0in 0in, scale=.45] {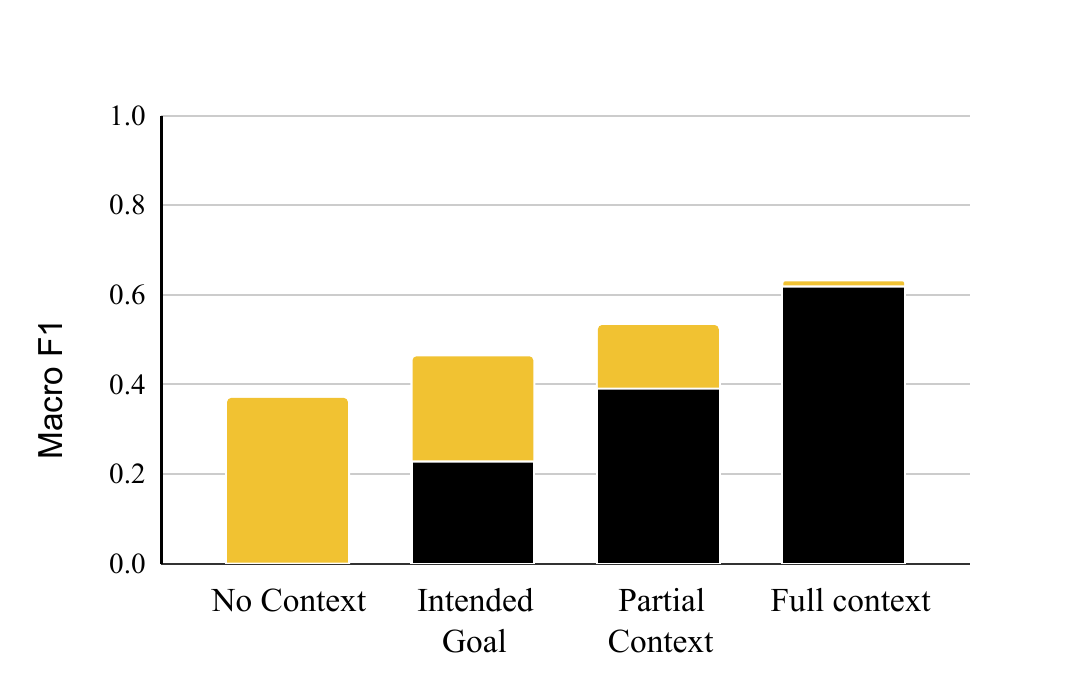}
	\caption{\textbf{Llama-70BI} is cautious when using partial contexts.}    
    \end{subfigure}
    \caption{Models are able to utilize conditions in addition to the story context to improve performance when identifying outcome labels. The black bars show performance with story context alone and the yellow bars show the improvement with annotated variant conditions used in conjunction with the story context. We show Llama models here and the other models in \cref{fig:fig7}. }
    \label{fig:fig3}
 \vspace{-5mm}
\end{figure}

\begin{figure}[t]
    \vspace{-5mm}
    \centering     
    \includegraphics[trim=0 0in 0 0in,  width=.9\columnwidth]
    {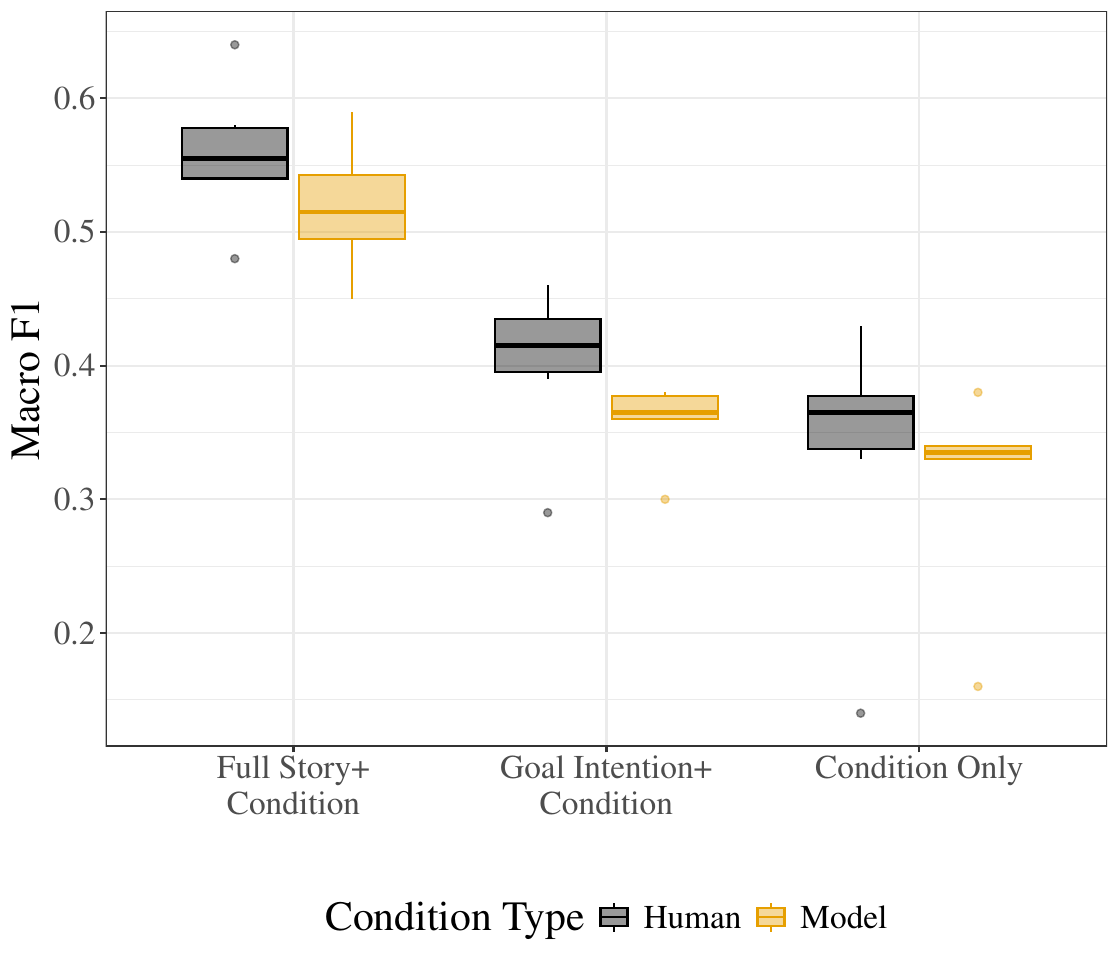}
    \vspace{-2mm}
	\caption{ \textbf{Model} generated conditions are not as good as \textbf{Human} annotated conditions leading to lower performance on outcome identification.  We   compare task settings with varying \textbf{SAGA story} contexts paired with an annotated or a generated condition. Results are aggregated across all models.} %

	\label{fig:fig4}
 \vspace{-5mm}
\end{figure}

\subsection{RQ3: When are conditions useful?} 
\label{sec:RQ3} 
We examine whether models are able to use conditions when available story context is incomplete. We identify if the outcome is true when given varying amounts of story context and a condition (C for crowd-sourced or GC for generated) as follows: 
\begin{enumerate}
[topsep=-5px,topsep=0px,noitemsep]
\item full story context (Full SC$ + $C, Full SC$+$GC), 
\item partial story context after removing the sentences from which the condition was derived (Par SC$+$C, Par SC$+$GC), 
\item indicating the participant's intent of achieving their goal (Goal Int$+$C, Goal Int$+$GC),
\item no story context (C, GC).  
\end{enumerate}
We use the baselines of only the story context (Full SC), the partial story context (Par SC) and the intended goal (Goal Int). See prompts in \cref{tab:rq3-prompts}. 

\paragraph{Evaluation:} We  compare the above settings using Macro F1 for True/False/Unsure labels.  

\paragraph{Results \& Discussion:}
We find that models are able to leverage the condition as shown in \cref{fig:fig3} and \cref{fig:fig7}. Performance improves when outcome-variant conditions are paired with the context. Some models are able to leverage additional information provided by implied conditions even when the full story context is available. Performance across all settings is better for  outcome-variant conditions (see \cref{tab:all-condition-outcome-labeling}).  
When the context is unconstrained, GPT-4o, Llama-70BI and Mistral-BI are more cautious (selecting `Unsure' instead of `Yes' or `No'  for the incomplete contexts).  FlanT5-XXL's higher performance is attributable to Story-Cloze (similar to our outcome identification task) being a part of  Flan's 1.8K fine-tuning tasks.

\paragraph{Influence of Generated Conditions}
We compare performance when using model generated conditions vs. all annotated conditions. Performance is reflective of a model's condition generation capability discussed in \cref{sec:RQ2-generation}. As seen in \cref{fig:fig4}, generated conditions lead to .03 - .05 drop in Macro F1 (averaged across models) for the different context settings. See \cref{apptab:generated-condition-outcome-labeling} for all models' performance.

\begin{figure}
    \includegraphics[trim=0 0in 0in 0in,  width=1\columnwidth] {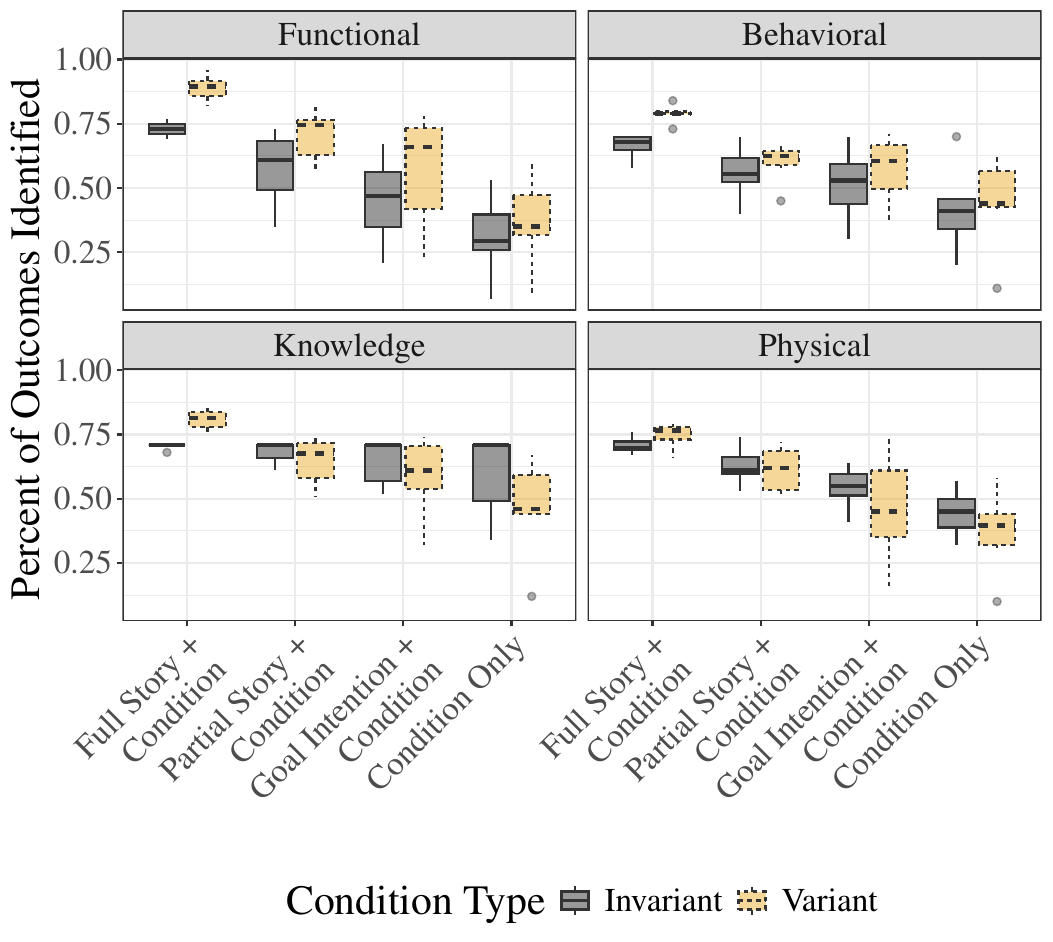}
    \vspace{-8mm}
    \caption{Models are better at utilizing functional conditions better than physical conditions for reasoning about outcome labels. We identify outcome labels using various story contexts and a variant or an invariant condition pair where we track the category of the first condition. Each boxplot aggregates over all six models.}
    \label{fig:fig5}
     \vspace{-5mm}
\end{figure}

\paragraph{Influence of Condition Category}
We examine which categories of conditions help models reason in both complete and incomplete story contexts. As seen in \cref{fig:fig5} performance across all groups is better for outcome- variant  than invariant conditions. For outcome-variant conditions, functional, knowledge and behavioral conditions lead to better performance than the physical when identifying outcome labels. The first condition in a condition pair determines the category and the pair determine the outcome relation.

\paragraph{Error Analysis:}
\label{sec:error-analysis} 
When a condition pair is not contrastive, or contrastive but not outcome variant, it can lead to poorer performance on outcome identification. This is worse in partial context settings where a model fails to fill in the missing context because it is unable to align the conditions with the context. We show examples of such misaligned conditions in \cref{tab:outcome-identification-errors}. 
Performance drop in incomplete context settings is mostly seen with Functional and Behavioral conditions than knowledge and physical categories (shown in \cref{fig:fig5}). We normalize outcome identification errors within each condition category.

 \begin{figure}[t]
    \centering     
    \includegraphics[trim=0 0in 0 0in,  width=.85\columnwidth]
    {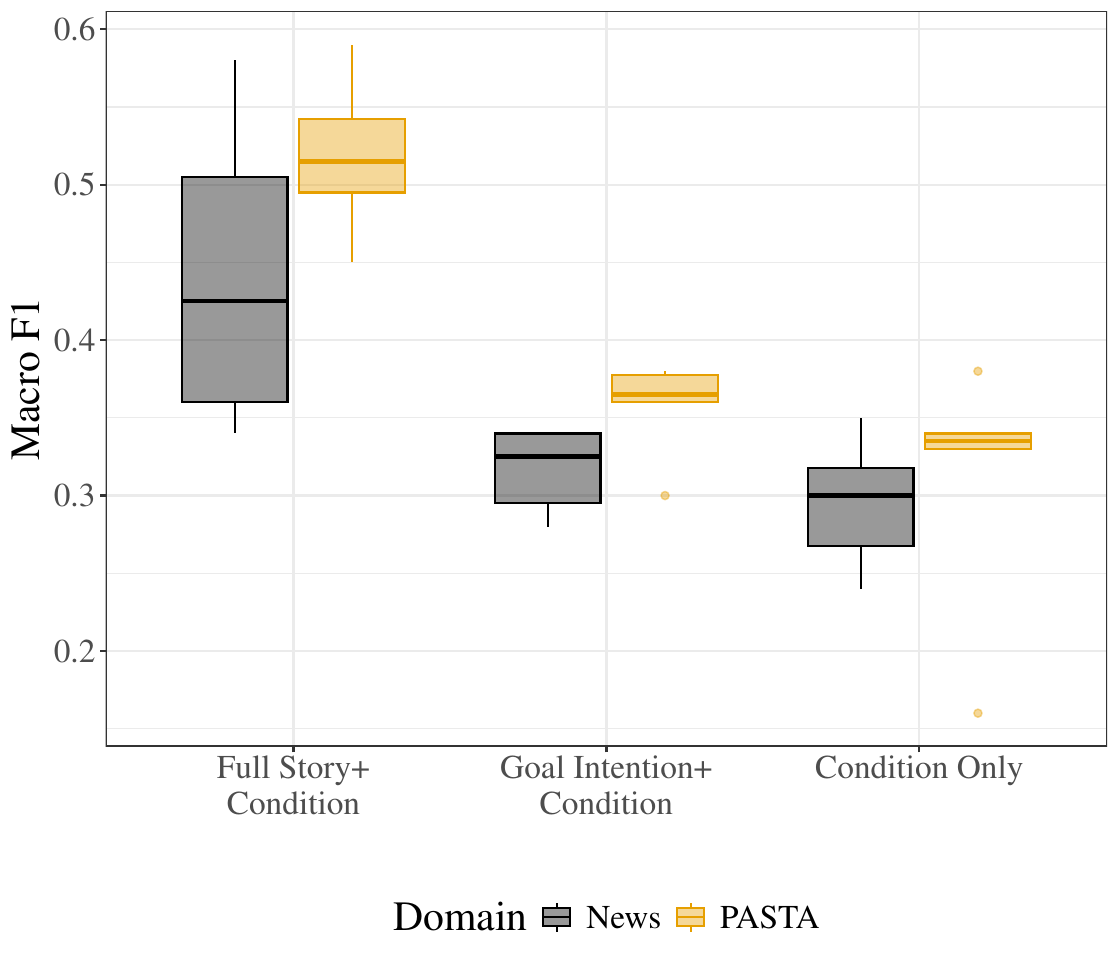}
    \vspace{-2mm}
	\caption{  We compare the performance of outcome identification for \textbf{PASTA/SAGA stories} and \textbf{SAGA News stories} using varying story contexts and generated conditions for both story types.  News stories are more complex leading to a drop in performance when using only the story context.  Conditions used with the story context fill some of the performance gap. Results are aggregated across all models.  %
 }
	\label{fig:fig6}
 \vspace{-3mm}
\end{figure}

\section{Cross-Domain Application to News}
\label{sec:news} 
We use the task prompts and setup from RQ2 and RQ3 to generate conditions for News stories and use them to identify outcome labels. News stories have longer sentences and include more complexities in the number of entities, topics, linguistic cues etc. when compared to SAGA stories. We examine their affect on model performance, comparing generated conditions from both story types.  
We compare the settings in RQ3, except for partial story context, as this requires identifying story sentence(s) supporting the condition.

\paragraph{Results \& Discussion:} The complexity of news increases the difficulty of the task and bias from models' knowledge of past events leads to a  performance drop of .11 when using the full story context (see \cref{apptab:news-generated-condition-outcome-labeling}).  When using conditions with the context, the drop is .02 -.08  as shown in \cref{fig:fig6}. 
See examples of model generations  in \cref{tab:news-examples}.

\section{Conclusions}
\label{sec:conclusions}
By defining \textit{Condition-based Reasoning tasks}, we have demonstrated the importance of identifying outcome-oriented relationships and their usefulness in validating story outcomes. To do this, we first combined and augmented two existing datasets to obtain outcome- variant and invariant conditions, and then showed that models can effectively use outcome-variant conditions in place of missing context.  We also showed how to generate and use outcome-variant conditions on a new domain. We examined and showed models' sensitivity to linguistic cues, which can indirectly affect both condition generation and outcome validation.  We hope that our work will help further linguistic examination of model behaviors to achieve a deeper and more robust narrative understanding.

\section{Limitations}
\label{sec:limit}
We acknowledge our work has the following limitations:
\begin{enumerate}
\item The linguistic analysis we performed is limited to the verbs used in our outcome descriptions.
\item The pre-trained large language models in our experiments can echo biases and mis-information either implicitly or explicitly. We do not attempt to control for these in this work.  
\item We focus on more formal written english and  our annotations are based on well known NLP data sources.  We use pre-trained large languages models to rewrite, generate and evaluate these annotations for use in our experiments.  
\item While we manually evaluated a subset of the generated data and did not find any misinformation, it is possible for the generated data to contain misinformation.
\item Our annotation and reasoning tasks do not examine biases in the data sources and hence do not control for them.  
\item Model generation and classification abilities can vary with the formality, style, and mood in the crowd written stories we annotated.
\end{enumerate}

\section*{Acknowledgments}
We wish to thank the anonymous reviewers for their helpful comments, feedback, and suggestions. %
We would also like to thank Katrin Erk, Sayontan Ghosh and Niranjan Balasubramian for early discussions and
feedback.
This material is based in part upon work supported by the National Science Foundation under Grant No. IIS-2024878. %
Some experiments were conducted on the UMBC HPCF, supported by the National Science Foundation under Grant No. CNS-1920079. %
This material is also based on research that is in part supported by the Army Research Laboratory, Grant No. W911NF2120076, and by DARPA for the SciFy program under agreement number HR00112520301. The U.S. Government is authorized to reproduce and distribute reprints for Governmental purposes notwithstanding any copyright notation thereon. The views and conclusions contained herein are those of the authors and should not be interpreted as necessarily representing the official policies or endorsements, either express or implied, of ARL, DARPA or the U.S. Government.

\bibliography{anthology,custom}
\bibliographystyle{acl_natbib}

\newpage
\clearpage
\appendix
\section{Appendix}
\label{sec:appendix}
\subsection{AI Assistance}
We did not use any AI assistants for writing  the paper or for any of the coding used in our experiments. All writing is original and entirely produced by the authors.

\subsection{Infrastructure} We used
both RTX 8000 with 48GB of GPU memory and
Nvidia A100 with 80GB GPU memory for inference with and without in-context examples.  Run time for a model ranges from a .5\-2 hours. 

\subsection{Hyperparameters}
We use a temperature of .6 for non-GPT models (1 for GPT models) and nucleus sampling to get the top 10\%. With these settings we were able to generate a variety of conditions that are closely related to the story. We obtain multiple  sufficiently diverse conditions through repeated sampling using these settings.  For simplicity, we use the same settings for all the tasks.

For in-context learning examples, we use a label balanced set of 2 examples for each of the Yes/No/Unsure when generating or identifying variant conditions.  Additionally, we only use alternate stories and their implicit conditions to ensure examples were unseen by the foundation models.   

\subsection{PASTA \& SAGA Data}
We provide examples of PASTA \& SAGA annotations we utilize  in \cref{tab:pasta-data-samples} and \cref{tab:saga-data-samples}.  We also use an SAGA News dataset of 250 goal annotations.

\subsection{Data Augmentation}
\label{app:data_augmentation}

We made the following augmentations to SAGA \& PASTA annotations for conditions and outcomes.  We provide annotation statistics in \cref{apptab:annotation-details} and annotation examples in \cref{tab:core-data-samples} and \cref{tab:core-news-data-samples}.

\begin{table}[t]
\centering
\resizebox{.98\columnwidth}{!}{
\small
\begin{tabular}[trim=0 .2in 0 0,  width=.98\linewidth]{p{0.48\linewidth} | p{0.48\linewidth}  }
\hline\hline
Annotation  &   Augmented SAGA/PASTA   \\ 
\hline\hline
Story Contexts & 886 (Actual) / 951 (Alternate) \\
\hline
Claims &  2985  \\
\# Condition Pairs  &  951 \\
\# Condition applied to Claim   & 6989\\
\hline
Claim Labels & 2125 / 299 / 561  \\ 
(identify claim truthfulness)&(True/False/Unknown) \\ 
\hline
Corrected Goal Labels  & 2125 / 484 / 376 \\
(identify goal achievement)&(Success/Unsuccess/Unsure) \\
\hline
Event-Goal Relations & 13140 / 1230 / 555 \\  
(5 relations per claim)& (Support/Oppose/No-Effect) \\ \hline
Condition Type & 6607 / 3448 / 3923\\ 
(condition-outcome relation)& (Support/Oppose/No-Effect)\\
\hline
Outcome-oriented Relations & 1985   / 694   / 598  \\
(a condition-pair's relation)& (variant/invariant/unsure)\\
\hline
\hline
\end{tabular}
}
\caption{ Augmented SAGA \& PASTA Data Statistics.  }
\label{apptab:annotation-details}
\vspace{-5mm}
\end{table}

\begin{table*}[t]
\centering
\resizebox{.98\textwidth}{!}{
\small
\begin{tabular}[trim=0 .2in 0 0,  width=.98\linewidth]{p{0.19\linewidth} | p{0.80\linewidth}  }
\hline\hline
PASTA Annotation & Annotated Data  \\ 
\hline\hline
\textbf{Original story 1} \newpage (story presented to annotators) &  Susan wanted to start a business. She decided to start an Italian restaurant in her home town. She hired a world famous chef to lead her kitchen. Her restaurant eventually became very successful. It was featured in many magazines and TV shows.  \\
\hline
Implied state \#1: &  Susan likes Italian food. \\
Supporting sentences: & 2\\

Preturbed state \#1: & Susan hates Italian food.\\
Supporting sentences: & 2 \\

Altered story supporting preturbed state \#1 \newpage \textbf{(Alternate Story \#1)}:& Susan wanted to start a business. She decided to start a Chinese restaurant in her home town. She hired a world famous chef to lead her kitchen. Her restaurant eventually became very successful. It was featured in many magazines and TV shows.\\
\hline
Implied state \#2: &The world famous chef knows Italian food very well.\\
Supporting sentences: & 2,4\\ 
Preturbed state \#2: & The world famous chef does not know Italian food at all.\\
Supporting sentences: & 4,5\\ 
Altered story supporting preturbed state \#2 \newpage \textbf{(Alternate Story \#2)}: &Susan wanted to start a business. She decided to start an Italian restaurant in her home town. She hired a world famous chef to lead her kitchen. Her restaurant was eventually a failure because the food was not Italian at all. She missed the chance to feature in many magazines and TV shows.\\
\hline\hline
\textbf{Original story 2} \newpage (story presented to annotators) &  Eli predicted the stock market trends on a lark. When every prediction came true his friends were in awe. They asked him to do it for the next day. His predictions turned out to be sheer luck. Eli’s friends were angry when
they lost money on their purchases.   \\
\hline
Implied state \#1: &  Eli does not know anything about the stock market. \\
Supporting sentences: & 1,4\\

Preturbed state \#1: & Eli is a highly talented stock analyst.\\
Supporting sentences: & 1,4,5 \\

Altered story supporting preturbed state \#1 \newpage \textbf{(Alternate Story \#1)} &Eli predicted the stock market trends as part of his job. When every prediction came true his friends were in awe. They asked him to do it for the next day. His predictions turned out to be ingenious. Eli's friends praised him when their purchases went up in value.\\
\hline\hline
\end{tabular}
}
\caption{Examples of \textbf{PASTA crowd annotations} showing two original stories. Each original story can have upto 3 pairs of Implied \& Preturbed state pairs (these examples have 2 and 1 pairs respectively).  Crowd workers infer an implied state and the story sentences supporting it.  They describe a counterfactual preturbed state and minimally alter the story sentences to support the preturbed state (altered sentences support the preturbed state).}
\label{tab:pasta-data-samples}
\vspace{-5mm}
\end{table*}

\begin{table*}[t]
\centering
\resizebox{.98\textwidth}{!}{
\small
\begin{tabular}[trim=0 .2in 0 0,  width=.98\linewidth]{p{0.19\linewidth} | p{0.80\linewidth}  }
\hline\hline
 Annotation & Data  \\ 
\hline\hline

\textbf{Original story 1} \newpage (presented to annotators) &  Susan wanted to start a business. She decided to start an Italian restaurant in her home town. She hired a world famous chef to lead her kitchen. Her restaurant eventually became very successful. It was featured in many magazines and TV shows.   \\
\hline
Volitional Participant: &  Susan\\
Susan’s Goal: &  Create a successful business\\
Goal Achievement in  &\\
\ \ \ Original story: &  Fully Successful\\
\ \ \ Alternate story \#1: & Fully Successful\\
\ \ \ Alternate story \#2: & Unsuccessful\\
Story Sentence \& relationship to Goal (obtained only for original story) & 1->Justifies some aspect of the goal \newpage 2->Enables or Helps to potentially achieve some aspect of the goal \newpage 3->Enables or Helps to potentially achieve some aspect of the goal \newpage 4->is an effect caused by an action related to the goal \newpage 5->Enables or Helps to potentially achieve some aspect of the goal\\
\hline\hline
\textbf{Original story 2} \newpage (presented to annotators) &   Eli predicted the stock market trends on a lark. When every prediction came true his friends were in awe. They asked him to do it for the next day. His predictions turned out to be sheer luck. Eli’s friends were angry when
they lost money on their purchases.   \\
\hline
Volitional Participant: &  Eli\\
Eli’s Goal: & Advise his friends on their finances.\\
Goal Achievement in  &\\
\ \ \ Original story: &  Unsuccessful\\
\ \ \ Alternate story \#1: & Fully Successful\\
Story Sentence \& relationship to Goal (obtained only for original story) & 1->is Related to another sentence, but, unrelated to the goal \newpage 2->Justifies some aspect of the goal \newpage 3->Justifies some aspect of the goal \newpage 4->Enables or Helps to potentially achieve some aspect of the goal \newpage 5->is an effect caused by an action related to the goal\\
\hline
Volitional Participant: &  Eli's friends\\
Eli’s friends's Goal: & to make some quick easy for sure money.\\
Goal Achievement in  &\\
\ \ \ Original story: &  Unsure\\
\ \ \ Alternate story \#1: & Fully Successful\\
Story Sentence \& relationship to Goal (obtained only for original story) &1->is Related to another sentence, but, unrelated to the goal \newpage 2->Justifies some aspect of the goal \newpage 3->Enables or Helps to potentially achieve some aspect of the goal \newpage 4->Prevents or Blocks the achievement of some aspect of the goal \newpage 5->is an effect caused by an action related to the goal\\
\hline\hline
\end{tabular}
}
\caption{Examples of \textbf{SAGA crowd annotations} showing two original stories. Each original story can have goals annotated for upto 4 participants (these examples have 1 and 2 participants respectively).  Crowd workers describe an overarching goal for a participant and identify whether the goal is achieved in the story and in all the alternate stories.  They also identify how each story sentence relates to the goal.}
\label{tab:saga-data-samples}
\vspace{-5mm}
\end{table*}

\begin{table*}[t]
\centering
\resizebox{.98\textwidth}{!}{
\small
\begin{tabular}[trim=0 .2in 0 0,  width=.98\linewidth]{p{0.45\linewidth} | p{0.54\linewidth}  }
\hline\hline
 Annotation & Data  \\ 
\hline\hline
\textbf{Original Story 1} & \\
Outcome described as an ongoing activity: & Susan was creating a successful business. \\
Outcome Label (Original story): & True \\
\hline
Outcome described as a completed activity: & Susan created a successful business. \\
Outcome Label (Original story): & True \\
Outcome Label (Alternate story 1): & True \\
Outcome Label (Alternate story 2): & False \\
\hline
For condition pair \#1 & \\
Story Sentence relationship for Implied state \#1: & Enables \\
Story Sentence relationship for Preturbed state \#1: & Opposes \\
Outcome Variance: & Outcome-Invariant \\
\hline
For condition pair \#2 & \\
condition-outcome relationship for Implied state \#2: & Enables \\

condition-outcome relationship for Preturbed state \#2: & Opposes \\

Outcome Relationship & Outcome-Variant \\
\hline\hline
\textbf{Original Story 2} & \\
Outcome described as an ongoing activity: & Eli was providing financial advice to his friends.\\
Outcome Label (Original story): & True \\
\hline
Outcome described as a completed activity: & Eli provided financial advice to his friends. \\
Outcome Label (Original story): & True\\
Outcome Label (Alternate story 1): & True \\
\hline
For condition pair \#1 & \\
condition-outcome relationship for Implied state \#1: & Enables \\
condition-outcome relationship for Preturbed state \#1: & Opposes \\
Outcome Relationship: & Outcome-Invariant \\
\hline
Outcome described as an ongoing activity: & Eli's friends were finding a quick and easy way to make some money.\\
Outcome Label (Original story): & False \\
\hline
Outcome described as a completed activity: & Eli's friends found a quick and easy way to make some money. \\
Outcome Label (Original story): & False\\
Outcome Label (Alternate story 1): & True \\
\hline
For condition pair \#1 & \\
condition-outcome relationship for Implied state \#1: & Opposes \\
condition-outcome relationship for Preturbed state \#1: & Enables \\
Outcome Relationship: & Outcome-Variant \\ 
\hline\hline
\end{tabular}
}
\caption{Examples of \textbf{our augmented annotations} showing the same two original stories from PASTA \& SAGA.  For outcome descriptions, we use GPT-3.5-Turbo to rewrite the goal descriptions from SAGA as complete sentences describing an ongoing and a completed activity.  The condition-outcome relationships are obtained automatically using the SAGA sentence relationships.  The outcome label for all outcomes is annotated by an expert based on the story and the outcome relationship is automatically identified based on these annotations.}
\label{tab:core-data-samples}
\vspace{-5mm}
\end{table*}

\begin{enumerate}
[(1)]
\item We rewrote the SAGA goal descriptions as ongoing and completed actions using GPT-3.5-Turbo.  SAGA goals are not always grammatically correct and complete sentences;  we wrote 3 seed examples (without using any story context) and with these rewrote 100 SAGA goals from the training data split as complete  grammatical sentences describing past completed  and ongoing actions.  The story context is not used for this rewrite, to ensure it is only a sentence rewrite without any story based reasoning.  The authors manually verified that the generated claims are grammatically correct  sentences reflecting a completed goal relevant to the story.  Using the verified examples we rewrote the remaining goals and manually verified using the same checks as with the first 100.
\item Asking whether the outcomes (both ongoing and completed actions) from the above step can be inferred from the story context (using both actual and alternate story contexts of SAGA), we manually annotated whether each outcome is true, false or unsure for the given story.  We computed agreement with the crowd annotations in step (4) to show that these manual annotations are in fairly good agreement with the crowd annotations.   
\item The goal achievement annotation in SAGA based on 5-point likert scale from Unsuccessful to Fully Successful is not always accurate.  We used the goal achievement evaluation scores from the 3 crowd workers for the test and validation data splits to automatically correct the achievement annotation 
\item We convert the 5-point likert scale to the 3 outcome labels of True/False/Unsure (Fully \& Moderately Successful labels are converted to an outcome label of true and Less \& Un- Successful labels are converted to an outcome label of false). We compute IAA between our outcome label annotations and these automatically corrected crowd annotations and achieve a Cohen's Kappa score of .64 (without the corrections this score is .54). A majority of disagreement is due to unsure labels and if we were to consider only the true and false labels the IAA is .73 (without the corrections this score is .63) showing that our claim label annotations are in good agreement  with the crowd, especially after applying corrections based on multiple worker evaluations.

\item Each of the 5 sentences in a SAGA actual story was assigned one of 6 labels to identify the relation between the event in the sentence and the goal by a crowd worker.  We map these 3 relation labels as follows to identify the relationship between the sentence and outcome: any goal justifying and enabling relations are labeled \textit{Support}, any goal blocking relation is labeled \textit{Oppose} and the remaining are labeled \textit{No-Effect}.  We realize this mapping can be noisy, especially for data in the training split which could not be corrected in the above step (the evaluation scores were only available for test and validation data splits).  Using the SAGA sentence relations we automatically identify one of the 3 above relations for each condition (using the PASTA sentence annotations that identify which sentences a condition is based upon).  We assume the counterfactual condition would have the opposite relation which can be noisy.  These relations identify how a condition relates to the outcome (aka outcome-oriented relation annotations).
\end{enumerate}

\begin{enumerate}[resume]
\item  We automatically identify if a pair of contrastive conditions are outcome- variant, invariant or unsure using the goal achievement annotations from the two stories.
\item  We use Spacy to obtain the root verb of PASTA states (aka conditions) and categorize these descriptions into stative and action-oriented expressions to identify the lexical type of the annotation.  About a third are activity oriented expressions and two-thirds are stative expressions. We also categorize the conditions using the 4 categories from \cref{sec:conditions-outcomes}. 
\item  We obtain a random 125 news stories from an extended version of the previously published SAGA \cite{vallurupalli-etal-2024-saga} annotation effort.  These news stories were selected randomly from the National, Foreign and Financial news desks of the Annotated New York Times (ANYT) newswire dataset to obtain stories that contain several participants involved in  situations reflecting the complexity of common everyday situations.

\item For the news stories, we followed the same process as in steps 1 and 2 above to obtain an outcome and its truth label.  

\item We do not annotate conditions, but obtain conditions using the condition generation task in \cref{sec:RQ2-generation}.

\end{enumerate}

\begin{table*}[t]
\centering
\resizebox{.98\textwidth}{!}{
\small
\begin{tabular}[trim=0 .2in 0 0,  width=.98\linewidth]{p{0.59\linewidth} | p{0.40\linewidth}  }
\hline\hline
 News Stories & Annotations   \\ 
\hline\hline

\textbf{Original Story 1:} It happens. Just when you think they are gone for good and gather the teddy bears, the little pillow and the tattered blankie and store them in that old trunk in the attic, one of them appears on the doorstep, ready to move back in - with her boyfriend. That' s more or less what happened the other day to Mary Hanford of Salisbury, N. C. Mrs. Hanford, 100, had every reason to figure that her daughter, Elizabeth, was long gone. After all, Elizabeth was 65 and married to a former senator from Kansas named Bob Dole, and they had been living in Washington (in the Watergate, yet) for the better part of 30 years. [National Desk] & \textbf{Annotations from the SAGA Extended Version:} \newpage 
Volitional Participant: Elizabeth \newpage 
Elizabeth's Goal: return to her childhood home.  \newpage 
\textbf{Our annotations:} \newpage
Outcome described as a completed activity:  Elizabeth returned to her childhood home. \newpage
Outcome Label (Original story):  True \\
\hline
\textbf{Original Story 2}: Bond prices for R. H. Macy \& Company weakened yesterday on Wall Street speculation that its merchandise shipments might be affected by new credit caution and by a ripple effect from other financially depressed companies. After the markets closed and in response to the widespread speculation about changing credit policies on Macy merchandise, Henry Kassebaum, a senior vice president in New York for the Heller Financial Corporation, a large company that finances merchandise shipments to retailers, said that the company had changed its supplier credit policy in regard to Macy from'' revolving credit'' to an'' order - by - order'' policy. In effect, this creates a tighter, more cautious position on granting credit on Macy shipments. Loss Reported Earlier. Macy declined to comment on the possibility of any restriction in its merchandise credit. [Financial Desk]& \textbf{Annotations from the SAGA Extended Version:} \newpage Participant: Macy \newpage Macy's goal: its merchandise shipments might be affected by new credit caution and by a ripple effect from other financially depressed companies. \newpage
\textbf{Our annotations} \newpage
Outcome described as a completed activity:   Macy's merchandise shipments were affected by new credit caution and by a ripple effect from other financially depressed companies. \newpage
Outcome Label (Original story):  True\\
\hline\hline
\end{tabular}
}
\caption{Examples of  \textbf{our augmented news annotations} with generated conditions and outcome identification. We follow a similar process to that of rewriting SAGA/PASTA annotations, using GPT-3.5-Turbo to rewrite the goal descriptions from SAGA (only describing a completed activity) and obtain expert annotations for outcome labels.  We generate conditions using the condition generation process in RQ2 (see \cref{sec:RQ2-generation}). }
\label{tab:core-news-data-samples}
\end{table*}

\begin{table*}[t]
\centering
\resizebox{.98\textwidth}{!}{
\small
\begin{tabular}[trim=0 .2in 0 0,  width=.98\linewidth]{p{0.12\linewidth} | p{0.86\linewidth}  }
\hline\hline
Type  &   Prompt   \\ 
\hline\hline
Direct & A:\{story\}  \\
Questions &B: \{outcome description of ongoing or completed action\} \\
&For the context in A, Is the statement in B true?  Please indicate with a 'Y' for yes, 'N' for no and 'U' for unsure.  Do not give an explanation.   \\
\hline
Imperfective & A: \{outcome description of past ongoing activity\} \\
Paradox &B: \{outcome description of past completed activity\} \\
(No context)&If the statement in A is true, does it necessarily mean that the statement in B is also true? Please indicate with a 'Y' for yes, 'N' for no and 'U' for unsure.  Do not give an explanation.\\
\hline
Imperfective & A: \{outcome description of past ongoing activity\} \\
Paradox &B: \{outcome description of past completed activity\} \\
(with story)&C: \{story\}
For the context in C, if the statement in A is true, does it necessarily mean that the statement in B is also true? Please indicate with a 'Y' for yes, 'N' for no and 'U' for unsure.  Do not give an explanation.\\
\hline
\hline
\end{tabular}
}
\caption{Prompts used in examining the truth value of outcomes in \textbf{RQ1} (\cref{sec:RQ1}).  }
\label{tab:rq1-prompts}
\end{table*}

\begin{table*}[t]
\centering
\resizebox{.98\textwidth}{!}{
\small
\begin{tabular}[trim=0 .2in 0 0,  width=.98\linewidth]{p{0.80\linewidth} | p{0.19\linewidth}  }
\hline\hline
Dowty's Imperfective Paradox & Inference \\ 
\hline\hline
\textbf{Prompt without context:} \newpage
\textbf{A:} Eli was making accurate predictions for the stock market. \newpage
\textbf{B:} Eli made accurate predictions for the stock market. \newpage 
If the statement in A is true, does it necessarily mean that the statement in B is also true? Please indicate with a 'Y' for yes, 'N' for no and 'U' for unsure.  Do not give an explanation.

\textbf{Prompt with story context:} \newpage
\textbf{A:} Eli was making accurate predictions for the stock market. \newpage
\textbf{B:} Eli made accurate predictions for the stock market. \newpage 
\textbf{C:} Eli predicted the stock market trends on a lark. When every prediction came true his friends were in awe. They asked him to do it for the next day. His predictions turned out to be sheer luck. Eli's friends were angry when they lost money on their purchases. \newpage 
For the context in C, if the statement in A is true, does it necessarily mean that the statement in B is also true? Please indicate with a 'Y' for yes, 'N' for no and 'U' for unsure.  Do not give an explanation.
& Gold Label: Y \newpage Generated Label:  \newpage \ \ without context: Y \newpage \ \ with story context: N  \\
\hline\hline
\textbf{Prompt without context:} \newpage
\textbf{A:} Cindy was adopting a puppy. \newpage 
\textbf{B:} Cindy adopted a puppy. \newpage 
 If the statement in A is true, does it necessarily mean that the statement in B is also true? Please indicate with a 'Y' for yes, 'N' for no and 'U' for unsure.  Do not give an explanation. \newpage
\textbf{Prompt with story context:} \newpage
\textbf{A:} Cindy was adopting a puppy. \newpage 
\textbf{B:} Cindy adopted a puppy. \newpage 
\textbf{C:} Cindy found a cute puppy advertised on facebook. She wanted the puppy so bad. Her husband decided to surprise her. He brought the puppy home to her. Cindy was so happy. \newpage 
For the context in C, if the statement in A is true, does it necessarily mean that the statement in B is also true? Please indicate with a 'Y' for yes, 'N' for no and 'U' for unsure.  Do not give an explanation.
& Gold Label: N \newpage Generated Label: \newpage \ \ without context: N \newpage \ \ with story context: Y\\
\hline\hline
\end{tabular}
}
\caption{Data examples to highlight findings from \textbf{RQ1} (\cref{sec:RQ1}).  The two prompts for the imperfective Paradox with and without the story context in \cref{tab:rq1-prompts} lead to different inferences when the GPT-4o is used to infer whether Dowty's Imperfective Paradox is true. In these examples, story context influences the model's pragmatic inference.}
\label{tab:rq1-examples}
\end{table*}

\begin{table*}[t]
\centering
\resizebox{.98\textwidth}{!}{
\small
\begin{tabular}[trim=0 .2in 0 0,  width=.98\linewidth]{p{0.80\linewidth} | p{0.19\linewidth}  }
\hline\hline
Direct Questions & Inference \\ 
\hline\hline
\textbf{A:}  Sam got a cold one day. He tried to ignore it. But it grew worse, so he went to the doctor. The doctor told Sam he had the flu! Sam had to take medicine and rest for a week. \newpage
\textbf{B:} Sam was healing from his cold., \newpage
For the context in A, Is the statement in B true? Please indicate with a ’Y’ for yes, ’N’ for no and ’U’ for
unsure. Do not give an explanation.
& Gold Label: Y \newpage Generated Label: N \\
\hline
\textbf{A:} Sam got a cold one day. He tried to ignore it. But it grew worse, so he went to the doctor. The doctor told Sam he had the flu! Sam had to take medicine and rest for a week. \newpage 
\textbf{B:} Sam healed from his cold. 
\newpage
For the context in A, Is the statement in B true? Please indicate with a ’Y’ for yes, ’N’ for no and ’U’ for
unsure. Do not give an explanation.
& Gold Label: Y \newpage Generated Label: N \\
\hline\hline
\textbf{A:} Jane cooked spinach and chicken for dinner. Her kids hated spinach. They refused to eat it. Jane promised them a new toy if they ate the spinach. Upon hearing this, her kids gobbled up the spinach! \newpage 
\textbf{B:}
Jane was encouraging her children to eat healthily. \newpage
For the context in A, Is the statement in B true? Please indicate with a ’Y’ for yes, ’N’ for no and ’U’ for
unsure. Do not give an explanation.
& Gold Label: Y \newpage Generated Label: N\\
\hline
\textbf{A:} Jane cooked spinach and chicken for dinner. Her kids hated spinach. They refused to eat it. Jane promised them a new toy if they ate the spinach. Upon hearing this, her kids gobbled up the spinach!\newpage 
\textbf{B:}
Jane encouraged her children to eat healthily. \newpage
For the context in A, Is the statement in B true? Please indicate with a ’Y’ for yes, ’N’ for no and ’U’ for
unsure. Do not give an explanation.
& Gold Label: Y \newpage Generated Label: N\\
\hline\hline
\end{tabular}
}
\caption{Data examples to highlight errors with direct questioning discussed in \textbf{RQ1} (\cref{sec:RQ1}). We prompt a model to infer the outcome of both the in-progress and completed action directly.  These examples show that the GPT-4o model with the direct prompt from \cref{tab:rq1-prompts} generates incorrect labels.}
\label{tab:rq1-examples-more}
\end{table*}

\begin{table*}[t]
\centering
\resizebox{.98\textwidth}{!}{
\small
\begin{tabular}[trim=0 .2in 0 0,  width=.98\linewidth]{p{0.12\linewidth} | p{0.86\linewidth}  }
\hline\hline
Type  &   Prompt   \\ 
\hline\hline
Condition  & A:\{story\}  \\
Generation&B: \{outcome\} \\
&Generate a pair of contrastive conditions relevant to the context in A and the statement in B.  Make sure the first condition is supported by the context in A.  The conditions should be stative expressions.  Do not describe activities.  \\
\hline\hline
Identify  & A:\{story\}  \\
Condition&B: \{outcome\} \\
Single Step&C: \{condition1\} \\
&D: \{condition2\} \\
&For the story in A, is the statement in B true for one of the conditions in C and D and false for the other condition?  Please indicate with a 'Y' for yes, 'N' for no and 'U' for unsure.  Do not give an explanation.   \\
\hline
Identify  & A:\{story\}  \\
Condition&B: \{outcome\} \\
Standard CoT&C: \{condition1\} \\
&D: \{condition2\} \\
&For the story in A, is the statement in B true for one of the conditions in C and D and false for the other condition?  Please indicate with a 'Y' for yes, 'N' for no and 'U' for unsure.  Do not give an explanation.  Lets think step by step.  \\
\hline
Identify  & A:\{story\}  \\
Condition&B: \{outcome\} \\
Alternate CoT&C: \{condition1\} \\
&D: \{condition2\} \\
&For the story in A, is the statement in B true for one of the conditions in C and D and false for the other condition?  Please indicate with a 'Y' for yes, 'N' for no and 'U' for unsure.  Do not give an explanation.  Lets think step by step.  We already know the condition in C {supports/does not support/is unrelated to} the outcome. The condition in D {supports/does not support/is unrelated to} the outcome.  Given this information the answer is \\
\hline
Identify  & A:\{story\}  \\
a Condition's&B:\{condition\} \\
Support for&C: \{outcome\} \\
the outcome&For the story in A, does the condition in B enable or support the statement in C?  Please indicate with a 'Y' for yes, 'N' for no and 'U' for unsure. Do not give an explanation. \\
\hline\hline
\end{tabular}
}
\caption{Prompts used for generating and identifying outcome variant conditions in \textbf{RQ2} (\cref{sec:RQ2}) The first prompt is for generating and the rest are for identifying. Generated outputs from the last prompt for the two conditions are provided as intermediate data for the Alternate CoT. }
\label{tab:rq2-prompts}
\vspace{-5mm}
\end{table*}

\begin{table*}[t]
\centering
\resizebox{.98\textwidth}{!}{
\small
\begin{tabular}[trim=0 .2in 0 0,  width=.98\linewidth]{p{0.36\linewidth} | p{0.63\linewidth} }
\hline\hline
Prompt & Model Generated Condition Pair \& any Issues with the pair  \\ 
\hline\hline
\textbf{A:} Anthony watches a lot of movies about computers. He wished he could impress a venture capitalist to invest in him. After college he designed a website. People were allowed to warn people about other people on this site. A venture capitalist saw the site, and told Anthony he was wasting his time. \newpage \textbf{B:} Anthony made money from his computer knowledge.  \newpage 
Generate a pair of contrastive conditions related to the context in A and the statement in B.  Make sure the first condition is supported by the context in A.  The conditions should be stative expressions.  Do not describe activities. & answer: Anthony was not able to impress a venture capitalist. \textbf{[FlanT5-XXL] Model does not generate condition pairs.} \newpage Anthony's website was not successful, Anthony's website was successful \textbf{[GPT-4o-mini]} \newpage Anthony's website was appealing to venture capitalists., Anthony's website was not appealing to venture capitalists. \textbf{[GPT-4o] The first condition is not supported by the story as expected.} \newpage Anthony was unsuccessful in making money from his computer knowledge, Anthony was successful in making money from his computer knowledge \textbf{[Mistral-7BI] conditions are good but lengthy.} \newpage The venture capitalist was impressed with Anthony's website, The venture capitalist was unimpressed with Anthony's website. \textbf{[Llama-8BI] The first condition is not supported by the story as expected.} \newpage Anthony is unsuccessful financially, Anthony is successful financially \textbf{[Llama-70BI]}\\
\hline
\textbf{A:} Sam had proposed to Nancy on the steps of the satanic church. Neither was interested in unduly waiting. The wedding date was set for Fall, only a month away. Nancy wore a cream colored dress and laced veil. Sam lifted the veil, kissing her and thinking of the coming honeymoon. \newpage \textbf{B:} Nancy married Sam as soon as possible. \newpage 
Generate a pair of contrastive conditions related to the context in A and the statement in B.  Make sure the first condition is supported by the context in A.  The conditions should be stative expressions.  Do not describe activities. & answer: The sand was not a good color. The sand was a good color. \textbf{[FlanT5-XXL] Unrelated to the story.} \newpage Nancy was not ready to marry Sam. Nancy was ready to marry Sam. \textbf{[GPT-4o-mini] The first condition is not supported by the story as expected.} \newpage The wedding planning was very efficient. Unexpected delays arose, postponing the wedding. \textbf{[GPT-4o] } \newpage A: The wedding date was not set for immediately., B: The wedding date was set for immediately. \textbf{[Mistral-7BI] The first condition is not supported by the story and uses list items.} \newpage Nancy is already engaged, Nancy is not married yet. \textbf{[Llama-8BI] The conditions are not contrastive.} \newpage The wedding was a month away., The wedding was immediate.  \textbf{[Llama-70BI] The first condition is not supported by the story} \\
\hline\hline
\end{tabular}
}
\caption{Examples of erroneous conditions generated by the various models for the task in \cref{sec:RQ2-generation}.}
\label{tab:rq2-generated-conditions-more}

\end{table*}

\begin{table*}[t]
\centering
\resizebox{.98\textwidth}{!}{
\small
\begin{tabular}[trim=0 .2in 0 0,  width=.98\linewidth]{p{0.14\linewidth} | p{0.85\linewidth}  }
\hline\hline
Type  &   Prompt   \\ 
\hline\hline
Full Context $+$  & A:\{story\}  \\
No Condition&B: \{outcome\}  \\
&For the context in A, Is the statement in B true?  Please indicate with a 'Y' for yes, 'N' for no and 'U' for unsure.  Do not give an explanation.     \\
\hline
Full Context $+$  & A:\{story\}  \\
Condition&B: \{condition\} \\
&C: \{outcome\}  \\
&For the context in A, and the condition in B, Is the statement in C true?  Please indicate with a 'Y' for yes, 'N' for no and 'U' for unsure.  Do not give an explanation.     \\
\hline
Partial Context $+$  & A:\{story without the sentences supporting the condition\}  \\
Condition&B: \{condition\} \\
&C: \{outcome\}  \\
&For the context in A, and the condition in B, Is the statement in C true?  Please indicate with a 'Y' for yes, 'N' for no and 'U' for unsure.  Do not give an explanation.     \\
\hline
Goal Intent $+$  & A:\{Participant\} wanted to achieve \{goal\}  \\
Condition&B: \{condition\} \\
&C: \{outcome\}  \\
&For the context in A, and the condition in B, Is the statement in C true?  Please indicate with a 'Y' for yes, 'N' for no and 'U' for unsure.  Do not give an explanation.     \\
\hline
\hline

\end{tabular}
}
\caption{Prompts used in examining the truth value of outcomes using conditions with varying context for \textbf{RQ3}(\cref{sec:RQ3}).  }
\label{tab:rq3-prompts}
\vspace{-5mm}
\end{table*}

\begin{table*}[t]
 \centering
\resizebox{.98\textwidth}{!}{
 \begin{tabular}{|l|l|c|c|c|c|c|c|c|}
 \hline
\multirow{2}{*}{Type}&&\multicolumn{2}{|c|}{Full Story Context}& \multicolumn{2}{|c|}{Partial Story Context} & \multicolumn{2}{|c|}{Intended Goal} & Only \\ 
&Model& Baseline \#1&$+$Condition&  Baseline \#2 & $+$Condition & Baseline \#3& $+$Condition &  Condition\\ 
\hline
\multirow{6}{*}{Var}&FlanT5-XXL  & .70 (.88/.78/.44) & \textbf{.60} (.89/.80/.12) &.39 (.68/.35/.13) & .53 (.79/.62/.18) &.30 (.74/.11/.05) & .50 (.80/.65/.06) & .47 (.59/.70/.12)\\
&GPT-4o-mini  & .66 (.87/.82/.29)& \textbf{.61} (.89/.84/.10) &.50 (.70/.64/.16) &.54 (.81/.70/.09) &.29 (.49/.26/.10) &.47 (.77/.61/.02) & .40 (.54/.55/.10)\\
&GPT-4o  & .67 (.84/.83/.33) & .66 (.86/.82/.30) &.35 (.42/.50/.13) &.48 (.62/.66/.15) &.05 (.02/.03/.09) & .29 (.40/.39/.12) & .13 (.07/.22/.09)\\
&Mistral-7BI  & .53 (.83/.76/.00)& \textbf{.59} (.83/.78/.18) & .42 (.60/.46/.19) &.49 (.64/.65/.18) & .30 (.67/.14/.09) &.40 (.53/.58/.10) & .37 (.41/.60/.09)\\
&Llama-8BI    & .53 (.83/.76/.00)& \textbf{.54} (.84/.76/.00) &.36 (.64/.42/.00) &.47 (.74/.68/.00) &.32 (.58/.39/.00) &.47 (.74/.68/.00) & .44 (.63/.63/.06)\\
&Llama-70BI    & .62 (.87/.79/.22)& \textbf{.63} (.87/.80/.22) & .39 (.55/.48/.12) &.53 (.67/.72/.19) & .23 (21/.46/.02) &.46 (.60/.69/.08) & .37 (.47/.57/.07)\\
\hline
\multirow{6}{*}{InVar}&FlanT5-XXL  & .52 (.82/.62/.12)& .50 (.82/.66/.03) &.47 (.75/.44/.21)&.47 (.77/.54/.12) &.28 (.74/.06/.03)&.43 (.72/.50/.08) & .39 (.43/.50/.25)\\
&GPT-4o-mini  & .58 (.81/.69/.23)& .53 (.80/.66/.14) &.43 (.62/.45/.23) &.46 (.74/.45/.19) &.33 (.54/.31/.15) &.41 (.66/.41/.15) & .33 (.42/.32/.26)\\
&GPT-4o  & .68 (.80/.72/.52)& .61 (.78/.62/.42) & .31 (.29/.36/.28) & .39 (.49/.40/.28) &.12 (.08/.03/.26) &.27 (.33/.22/.26) & .15 (.07/.11/.26)\\
&Mistral-7BI  & .49 (.77/.63/.06)& .51 (.76/.61/.15) & .43 (.72/.51/.05) &.45 (.64/.47/.24) & .37 (.69/.23/.20) & .36 (.43/.38/.27) & .30 (.29/.39/.24) \\
&Llama-8BI    & .49 (.81/.66/.00) & .44 (.73/.59/.00) & .41 (.72/.51/.00) &.41 (.68/.54/.00) & .34 (.61/.40/.00) &.37 (.63/.47/.00) & .34 (.53/.46/.04)\\
&Llama-70BI    & .54 (.78/.63/.20)& .54  (.77/.64/.21) &.46 (.71/.51/.16) &.48 (.69/.59/.16) &.22 (.18/.44/.04) & .38 (.46/.46/.21) & .33 (.33/.37/.29)\\
\hline 
\multirow{6}{*}{All}&FlanT5-XXL & .58 (.85/.69/.19)& .54 (.85/.72/.05) &.43 (.72/.40/.18)&.49 (.77/.57/.14) &.29 (.74/.09/.03)&.46 (.75/.56/.08) &.43 (.50/.50/.21)\\
&GPT-4o-mini  & .61 (.83/.75/.24)& .57 (.84/.74/.13) &.47 (.65/.54/.21)&.50 (.77/.57/.16) &.31 (.52/.29/.13)&.44 (.70/.50/.11) & .37 (.42/.43/.21)\\
&GPT-4o       & \textbf{.69} (.81/.77/.47)& .64 (.81/.71/.39) &.33 (.34/.43/.23)& .44 (.55/.53/.24) & .09 (.05/.03/.19)& .29 (.36/.30/.21) & .14 (.07/.16/.20)\\
&Mistral-7BI  & .51 (.80/.69/.05)& .54 (.78/.68/.16) &.42 (.68/.49/.10)&.47 (.64/.55/.22) &.34 (.68/.19/.16)&.39 (.47/.47/.22) & .33 (.34/.48/.19)\\
&Llama-8BI    & .51 (.82/.70/.00)& .48 (.78/.66/.00) &.39 (.69/.47/.00)&.43 (.70/.60/.00) & .33 (.60/.40/.00)&.41 (.67/.56/.00) & .38 (.57/.53/.05)\\
&Llama-70BI   & .57 (.81/.70/.21)& .58  (.81/.71/.22) & .43 (.65/.50/.14)& .50 (.68/.65/.17) & .22 (.19/.45/.03) &.42 .52/.57/.17) & .36 (.39/.46/.22)\\
\hline 
\end{tabular}
}
 \caption{Outcome validation performance (Macro F1 and individual F1 scores of True/False/Unsure labels when \textbf{using outcome-variant \& invariant conditions}.  The 'All' type shows F1 when both condition types are used together. We compare the 4 varying contexts listed in \cref{sec:RQ3} and 3 baselines consisting of only the story contexts -- full, partial and goal. \textbf{Bolded} numbers show variant conditions lead to 2-8 \% improvement in model performance (except for GPT-4o \& GPT-4o-mini) over the 'All' baseline \#1. }
\label{tab:all-condition-outcome-labeling}
\vspace{-3mm}
\end{table*}

\begin{table*}[t]
 \centering
\resizebox{.98\textwidth}{!}{
 \begin{tabular}{|l|c|c|c|c|}
 \hline
&Full Story & \multicolumn{2}{|c|}{Varying Context With Condition }  & Only \\ 
Model  & Baseline&Full Story & Intended Goal&  Condition\\ 
\hline
FlanT5-XXL  & .57 (.82/.71/.19)& .55 (.80/.69/.14) & .38 (.66/.38/.09) & .33 (.45/.36/.18) \\
GPT-4o-mini  & .60 (.81/.77/.21)& .51 (.74/.70/.10) & .36 (.59/.37/.11) &.34 (.41/.38/.23) \\
GPT-4o  & .69 (.81/.79/.46)& .59 (.72/.70/.35) & .30 (.39/.32/.20) &.16 (.10/.19/.19) \\
Mistral-7BI    & .52 (.76/.71/.09)& .49 (.69/.66/.12) & .38 (.39/.53/.21) & .33 (.36/.53/.09) \\
Llama-8BI    & .49 (.77/.70/.00)& .45 (.68/.66/.00) &.36 (.55/.52/.00) &.34 (.51/.52/.00) \\
Llama-70BI    & .55 (.78/.73/.13)&.52 (.73/.69/.15) &.37(.42/.57/.13) &.38 (.41/.54/.19) \\
\hline 
\end{tabular}
}
 \caption{Outcome validation performance (F1 scores of True/False/Unsure labels) \textbf{using generated conditions}.  We compare the 4 varying contexts leaving out the Partial Story context as we do not identify which sentences the condition depends upon. see \cref{sec:RQ3} for the task setup. We could not generate conditions for some stories and performance with generated conditions is lower than with annotated conditions for all models.     }
\label{apptab:generated-condition-outcome-labeling}
\vspace{-3mm}
\end{table*}

\begin{table*}[t]
\centering
\resizebox{.98\textwidth}{!}{
\small
\begin{tabular}[trim=0 .2in 0 0,  width=.98\linewidth]{p{0.49\linewidth} | p{0.49\linewidth} }
\hline\hline
 Prompt &  Model generations \& Issues\\ 
\hline\hline

\textbf{A:} The proposal was unsuccessful. \newpage
\textbf{B:} Danny proposed to Beth. \newpage
For the condition in A, is the statement in B true? \newpage Please indicate with a 'Y' for yes, 'N' for no and 'U' for unsure. Do not give an explanation. 
& \textbf{FlanT5-XXL} incorrectly generates an N when only the condition is provided but correctly generates a Y when story context is availablen as in the next row. \newpage \textbf{GPT-4o-mini} correctly generates Y for all settings where the condition is provided and incorrectly generates a N otherwise \newpage \textbf{GPT-4o} generates a U for all contexts where the condition is provided and correctly generates a Y with story context. \newpage \textbf{Mistral7BI} incorrectly generates a N when the condition is provided but correctly generates a Y with story context. \newpage \textbf{Llama8BI} correctly generates a Y for all settings where the condition is provided. \newpage \textbf{Llama70BI} incorrectly generates N because the condition states something that happened later.  For the settings where some story context is available the model correctly generates Y.\\
\hline\hline
\textbf{A:} Danny got down on one knee. He asked Beth to marry him. Beth felt very awkward. She didn't want to marry Danny! To be honest, she told him she did not want to marry him. \newpage
\textbf{B:} Beth was brutally honest. \newpage
\textbf{C:} Beth let Danny down gently. \newpage
For the context in A and the condition in 
B, Is the statement in C true?  \newpage Please indicate with a 'Y' for yes, 'N' for no and 'U' for unsure. Do not give an explanation. & \textbf{FlanT5-XXL} correctly generates a Y when provided with story context and generates a N or U otherwise. \newpage  \textbf{GPT-4o-mini} correctly generates a Y for all settings where the story context is provided and generates a U otherwise. \newpage \textbf{GPT-4o} correctly generates a Y for all settings where the story context is provided and generates a U otherwise. \newpage \textbf{Mistral7BI} correctly generates a Y for all settings where the story context is provided and incorrectly generates a N otherwise. \newpage \textbf{Llama8BI} correctly generates Y and generates N for all settings where the condition is provided. \newpage \textbf{Llama70BI} correctly generates Y whereas it incorrectly generates N when only the context is provided.\\
\hline\hline
\end{tabular}
}
\caption{Examples of \textbf{outcome identification errors} from the various settings in \cref{sec:RQ3} using Model generated conditions. We show 2 prompts with one not using a story context and the other using a story context.  We show how models behave for various settings using a single story and 2 outcome descriptions for the two participants.  }
\label{tab:outcome-identification-errors}
\vspace{-5mm}
\end{table*}

\begin{table*}[t]
\centering
\small
\resizebox{.98\textwidth}{!}{
\begin{tabular}[trim=0 .2in 0 0,  width=.98\linewidth]{p{0.49\linewidth} | p{0.49\linewidth}   }
\hline\hline
 Prompt & Model Generations \& Issues\\ 
\hline\hline
\textbf{A:} When George Bush declared last June that he would not raise taxes, skeptical reporters gave him repeated chances to hedge. Could he support higher excise taxes ?  No, Mr. Bush said. Would he use'' revenue enhancers'' ?  No, Mr. Bush insisted. \newpage
\textbf{B: } George Herbert Walker Bush did not raise taxes \newpage
Generate a pair of contrastive conditions related to the context in A and the statement in B.  Make sure the first condition is supported by the context in A.  The conditions should be stative expressions.  Do not describe activities. & answer: George Bush raised taxes. \textbf{[FlanT5-XXL]} \newpage 
George Bush is evasive., George Bush is straightforward. \textbf{[ GPT-4o-mini]} \newpage
George Bush is committed to his promise., George Bush is open to changing his stance.\textbf{[ GPT-4o]} \newpage
George Herbert Walker Bush did not commit to not raising taxes., George Herbert Walker Bush committed to not raising taxes.\textbf{[ Mistral-7BI]} \newpage
George Bush had a plan to increase taxes.,  George Bush did not have a plan to increase taxes.\textbf{[ Llama-8BI]} \newpage
George Bush was open to new taxes.  George Bush was not open to new taxes.
\textbf{[ Llama-70BI]}\\
\hline
\textbf{A:} When George Bush declared last June that he would not raise taxes, skeptical reporters gave him repeated chances to hedge. Could he support higher excise taxes ?  No, Mr. Bush said. Would he use ``revenue enhancers'' ?  No, Mr. Bush insisted. \newpage
\textbf{B:} \{generated condition (the first one) from the above row\} \newpage 
\textbf{C: } George Herbert Walker Bush did not raise taxes \newpage
For the context in A and the condition in B, Is the statement in C true? \newpage Please indicate with a 'Y' for yes, 'N' for no and 'U' for unsure. Do not give an explanation.
& \textbf{FlanT5-XXL} generates correctly a Y for all contexts containing the story or the George Bush's intended goal and generate N for the condition only setting. \newpage
\textbf{GPT-4o-mini} correctly generates a Y when the condition is not part of the context but generates a N when it is. \newpage \textbf{GPT-4o} generates a U for most contexts and a N when the condition provides the most context. \newpage \textbf{Mistral-7BI} correctly generates a Y when the condition is not part of the context but generates a N when it is.\newpage \textbf{Llama8BI} incorrectly generates a N for all contexts using its prior knowledge \newpage \textbf{Llama70BI} incorrectly generates a N for all contexts using its prior knowledge  \\
\hline\hline
\end{tabular}
}
\caption{Examples of \textbf{News story}  condition and outcome generations which are \textbf{biased due to models' knowledge of past news events}. All Models except FlanT5-XXL and Mistral make use of their knowledge of prior events when generating conditions and identifying outcome labels. FlanT5-XXL  does not generate condition pairs but uses its available context; Mistral-7BI generates correct but long condition pairs and considers all available context. }
\label{tab:news-examples}
\vspace{-5mm}
\end{table*}

\begin{table*}[t]
 \centering
\resizebox{.98\textwidth}{!}{
 \begin{tabular}{|l|c|c|c|c|}
 \hline
 &Full Story & \multicolumn{2}{|c|}{Varying Context With Condition }  &  \\ 
Model& Baseline & Full Story & Intended Goal&  Only Condition\\ 
\hline
FlanT5-XXL  & .54 (.78/.44/.38)& .58 (.74/.41/.40) & .34 (.63/.20/.19) & .35 (.42/.31/.32) \\
GPT-4o-mini  & .51 (.75/.50/.28)& .46 (.72/.42/.23) &.29 (.54/.30/.05) & .32 (.45/.32/.21) \\
GPT-4o  & .60 (.68/.55/.56)& .58 (.69/.52/.53) & .34 (34/.33/.35)  &.31 (.27/.32/.33) \\
Mistral-7BI  & .40 (.74/.44/.03)& .35 (.54/.40/.12) & .34 (.33/.38/.30) &.24 (.19/.35/.18) \\
Llama-8BI    & .39 (.73/.44/.00)& .39 (.71/.45/.00) &.31 (56/.37/.00) &.26 (.44/.34/.00) \\
Llama-70BI   & .37 (.64/.40/.05)& .34 (.60/.41/.03) &.28 (.46/.39/.00) &.29 (.43/.38/.06) \\

\hline 
\end{tabular}
}
 \caption{\textbf{News Story} Outcome validation performance (Macro F1 and F1 scores of True/False/Unsure labels in parenthesis) \textbf{using generated conditions}.  We compare the 4 varying contexts leaving out the Partial Story context as we do not identify which sentences the condition depends upon. See \cref{sec:news} for the task details. }
\label{apptab:news-generated-condition-outcome-labeling}
\vspace{-3mm}
\end{table*}
\begin{figure*}[t] 
    \centering  
    \centering     
    \includegraphics[trim=0 0in 0 0in,  width=1\textwidth]
    {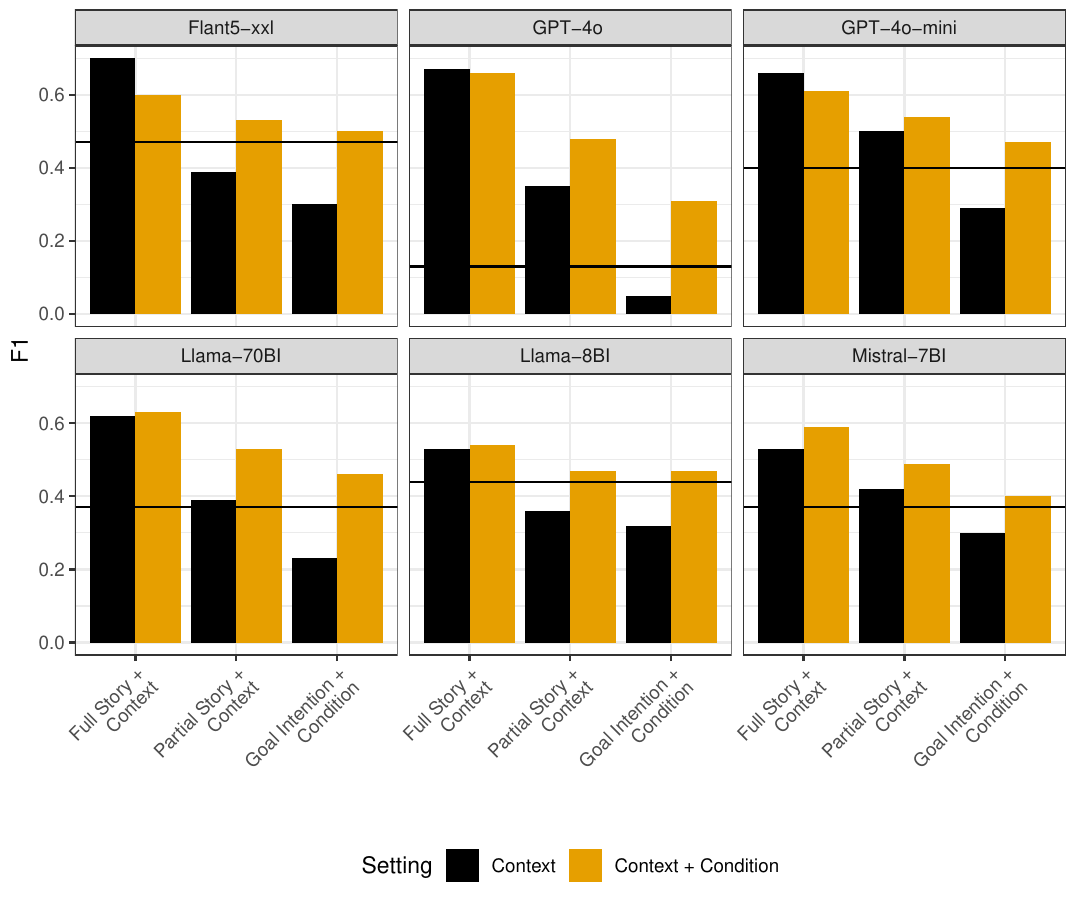}

    \caption{ Models are able to utilize conditions in addition to the story context to improve performance when identifying outcome labels. The exceptions are FlanT5-XXL, GPT-4o-mini and GPT-4o which drop in performance when using conditions with full story context. The black bar shows performance (Macro F1) with story context alone and the yellow bar shows the improvement when annotated variant conditions are used in addition to the story contexts. The black line indicates the performance of the condition only task setting. }
    \label{fig:fig7}
\end{figure*}

\end{document}